\documentclass[final]{cvpr}
\usepackage{times}
\usepackage{epsfig}
\usepackage{ifthen}
\usepackage{graphicx}
\usepackage{amsmath}
\usepackage{mathtools}
\usepackage{amssymb}
\usepackage{subcaption}
\usepackage{wrapfig}
\usepackage{multirow}
\usepackage{parskip}
\usepackage{array}
\usepackage{enumitem}       
\usepackage{algorithm2e,setspace}
\usepackage{changepage}
\usepackage{amsthm}
\usepackage[document]{ragged2e}
\justifying

\usepackage{xcolor}
\usepackage[resetlabels,labeled]{multibib}
\newcites{S}{Supplement References}
\usepackage[pagebackref=true,breaklinks=true,colorlinks,bookmarks=false]{hyperref}

\usepackage{fancyhdr}

\pagecolor{white}

\newcommand{\citesupp}[1]{\citeS{supp#1}}

\newcommand{\colornum}[1]{{\textbf{\color[RGB]{39, 158, 206}#1}}}
\newcommand{\mat}[1]{\MakeUppercase{\mathbf{#1}}}
\newcommand{\putfig}[1]{}
\renewcommand{\vec}[1]{\MakeLowercase{\mathbf{#1}}}
\newcommand{\tmat}[1]{$\mat{#1}$}

\newcommand{\myparagraph}[2][-.1]{\vspace{#1em}\noindent{\bf #2}}

\begin{document}
\justifying

\title{Convolutional Dynamic Alignment Networks for Interpretable Classifications}

\author{Moritz Böhle\\
\normalsize{MPI for Informatics}\\ 
\normalsize{Saarland Informatics Campus}
\and
Mario Fritz\\
\normalsize{CISPA Helmholtz Center}\\
\normalsize{for Information Security}
\and
Bernt Schiele\\
\normalsize{MPI for Informatics}\\ 
\normalsize{Saarland Informatics Campus}
}

\twocolumn[{%
\renewcommand\twocolumn[1][]{#1}%
\maketitle
\begin{center}
    \centering
    \captionsetup{type=figure}
    \includegraphics[width=.925\textwidth]{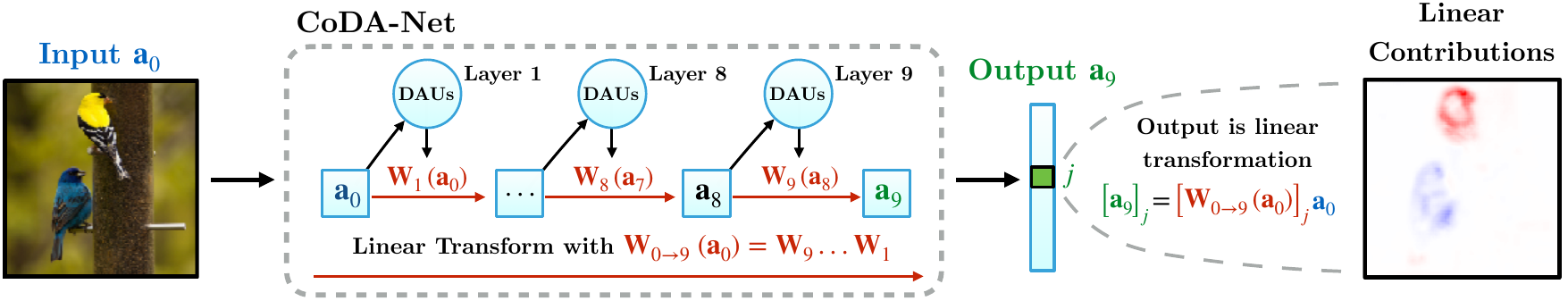}
    \captionof{figure}{\small {
    Sketch of a 9-layer CoDA-Net, which computes its {\bf \color[RGB]{0, 136, 43} output $\vec {a_9}$} for an  {\bf \color[RGB]{3, 101, 192}input $\vec {a_0}$} as a linear transform via a  matrix {\color[RGB]{200, 37, 6}$\mat {W_{0\rightarrow9}(\vec a_0)}$}, such that the output can be linearly decomposed into input contributions (see right). {\color[RGB]{200, 37, 6}$\mat {w_{0\rightarrow9}}$} is computed successively via multiple layers of Dynamic Alignment Units (DAUs), which produce matrices {\color[RGB]{200, 37, 6}$\mat w_l$} that align with their respective inputs $\vec a_{l-1}$. As a result, the combined matrix {\color[RGB]{200, 37, 6}$\mat {w_{0\rightarrow9}}$} aligns well with task-relevant patterns. Positive (negative) contributions for the class `goldfinch' are shown in red (blue).}}
    \label{fig:teaser}
\end{center}%
\bigskip
}]
\thispagestyle{fancy}


\begin{abstract}
We introduce a new family of neural network models called Convolutional Dynamic Alignment Networks\footnote{Code will be available at \url{github.com/moboehle/CoDA-Nets}.} (CoDA-Nets),
which are performant classifiers with a high degree of inherent interpretability. 
{Their core building blocks are Dynamic Alignment Units (DAUs), which \mbox{{linearly transform}} their input with weight vectors that \mbox{{dynamically align}} with task-relevant patterns.} As a result, CoDA-Nets model the classification prediction through a series of input-dependent linear transformations, allowing for linear decomposition of the output into individual input contributions.
    Given the alignment of the DAUs, the resulting contribution maps 
    align with discriminative input patterns. 
These model-inherent decompositions are of high visual quality and  outperform existing attribution methods under quantitative metrics. 
Further, CoDA-Nets constitute performant classifiers, achieving on par results 
to ResNet and VGG models on e.g. CIFAR-10 and TinyImagenet.

\end{abstract}
%
%
\section{Introduction}
\label{sec:intro}

Neural networks are powerful models that excel at a wide range of tasks.
However, they are notoriously difficult to interpret and extracting explanations 
    for their predictions is an open research problem. Linear models, in contrast, are generally considered interpretable, because
    the \emph{contribution} 
    (`the weighted input') of every dimension to the output is explicitly given.
Interestingly, many modern neural networks implicitly model the output as a linear transformation of the input;
    a ReLU-based~\cite{nair2010rectified} neural network, e.g.,
    is piece-wise linear and the output thus a linear transformation of the input, cf.~\cite{montufar2014number}.
    However, due to the highly non-linear manner in which these linear transformations are `chosen', the corresponding contributions per input dimension do not seem to represent the learnt model parameters well, cf.~\cite{adebayo2018sanity}, and a lot of research is being conducted to find better explanations for the decisions of such neural networks, cf.~\cite{simonyan2013deep,springenberg2014striving,zhou2016CAM,selvaraju2017grad,shrikumar2017deeplift,sundararajan2017axiomatic,srinivas2019full,bach2015pixel}.
    
In this work, we introduce a novel network architecture, the \textbf{Convolutional Dynamic Alignment Networks (CoDA-Nets)}, {for which the model-inherent contribution maps are faithful projections of the internal computations and thus good `explanations' of the model prediction.} 
There are two main components to the interpretability of the CoDA-Nets. 
    First, the CoDA-Nets are \textbf{dynamic linear}, i.e., they compute their outputs through a series of input-dependent linear transforms, which are based on our novel \mbox{\textbf{Dynamic Alignment Units (DAUs)}}. 
        As in linear models, the output can thus be decomposed into individual input contributions, see Fig.~\ref{fig:teaser}.
    Second, the DAUs are structurally biased to compute weight vectors that \textbf{align with \mbox{relevant} patterns} in their inputs. 
In combination, the CoDA-Nets thus inherently  
produce contribution maps that are `optimised for interpretability': 
since each linear transformation matrix and thus their combination is optimised to align with discriminative features, the contribution maps reflect the most discriminative features \emph{as used by the model}.

With this work, we present a new direction for building inherently more interpretable neural network architectures with high modelling capacity.
In detail, we would like to highlight the following contributions:
\begin{enumerate}[wide, label={\textbf{(\arabic*)}}, itemsep=-.5em, topsep=0em, labelwidth=0em, labelindent=0pt]
    \item We introduce the Dynamic Alignment Units (DAUs), which 
    improve the interpretability of neural networks and have two key properties:
    they are 
    \emph{dynamic linear} 
    and align their weights with discriminative input patterns.
    \item Further, we show that networks of DAUs \emph{inherit} these two properties. In particular, we introduce Convolutional Dynamic Alignment Networks (CoDA-Nets), which are built out of multiple layers of DAUs. As a result, the \emph{model-inherent contribution maps} of CoDA-Nets highlight discriminative patterns in the input.
    \item We further show that the alignment of the DAUs can be promoted 
    by applying a `temperature scaling' to the final output of the CoDA-Nets. 
    \item We show that the resulting contribution maps 
    perform well under commonly employed \emph{quantitative} criteria for attribution methods. Moreover, under \emph{qualitative} inspection, we note that they exhibit a high degree of detail.
    \item Beyond interpretability, 
    CoDA-Nets are performant classifiers and yield competitive classification accuracies on the CIFAR-10 and TinyImagenet datasets.
\end{enumerate}
\section{Related work}
\label{sec:related}

\myparagraph[0]{Interpretability.}
In order to make machine learning models more interpretable, a variety of techniques has been developed.
On the one hand, 
    research has been undertaken to develop model-agnostic explanation methods for which the model behaviour
    under different inputs is analysed; this includes among others \cite{lundberg2017unified,petsiuk2018rise,lime}.
    While their generality and the applicability to any model are advantageous,
    these methods typically require evaluating the respective model several times and are therefore costly
    approximations of model behaviour.
On the other hand,
    many techniques that explicitly take advantage of the internal computations have been proposed for explaining
    the model predictions, including, for example, \cite{simonyan2013deep,springenberg2014striving,zhou2016CAM,selvaraju2017grad,shrikumar2017deeplift,sundararajan2017axiomatic,srinivas2019full,bach2015pixel}.\\
In contrast to techniques that aim to explain models \emph{post-hoc},
some recent work has focused on designing new types of network architectures, which are \emph{inherently} more interpretable.
Examples of this are the prototype-based neural networks~\cite{chen2019looks}, the BagNet~\cite{brendel2018approximating}
and the self-explaining neural networks (SENNs)~\cite{melis2018towards}.
Similarly to our proposed architectures,
    the SENNs and the BagNets derive their explanations 
    from a linear decomposition of the output into contributions from the input (features).
This \emph{dynamic linearity}, i.e., the property that the output is computed via some form of an input-dependent linear mapping, is additionally shared by the entire model family of piece-wise linear networks (e.g., ReLU-based networks). In fact, the contribution maps of the CoDA-Nets are conceptually similar to  evaluating the `Input$\times$Gradient' (IxG), cf.~\cite{adebayo2018sanity}, on piece-wise linear models, which also yields a linear decomposition in form of a contribution map.
However, in contrast to the piece-wise linear functions, we combine this \emph{dynamic linearity} with a structural bias towards an alignment between the contribution maps and the discriminative patterns in the input. This results in explanations of much higher quality, whereas IxG on piece-wise linear models has been found to yield unsatisfactory explanations of model behaviour~\cite{adebayo2018sanity}.

\myparagraph{Architectural similarities.} In our CoDA-Nets, the convolutional kernels are dependent on the specific patch that they are applied on; i.e., a CoDA-Layer might apply different filters at every position in the input. As such, these layers can be regarded as an instance of dynamic local filtering layers as introduced in~\cite{jia2016dynamic}.
Further, our dynamic alignment units (DAUs) share some high-level similarities to attention networks, cf.~\cite{xu2015show}, in the sense that each DAU has a limited budget to distribute over its dynamic weight vectors (bounded norm), which is then used to compute a weighted sum. However, whereas in attention networks the weighted sum is typically computed over vectors, which might even differ from the input to the attention module, a DAU outputs a \emph{scalar} which is a weighted sum of all scalar entries in the input. Moreover, we note that at their optimum (maximal average output over a set of inputs), the DAUs solve a constrained low-rank matrix approximation problem~\cite{eckart1936approximation}. While low-rank approximations have been used for increasing parameter efficiency in neural networks, cf.~\cite{yu2017compressing}, this concept has to the best of our knowledge not been used in order to endow neural networks with a structural bias towards finding low-rank approximations of the input for increased interpretability in classification tasks. Lastly, the CoDA-Nets 
are related to capsule networks. However, whereas in classical capsule networks the activation vectors of the capsules directly serve as input to the next layer, in CoDA-Nets the corresponding vectors are used as convolutional filters. 
We include a detailed comparison in the supplement. 
\section{Dynamic Alignment Networks}
\label{subsec:alignment}

In this section, we present our novel type of network architecture: the Convolutional Dynamic Alignment Networks (CoDA-Nets). For this, we first introduce Dynamic Alignment Units (DAUs) as the basic building blocks of CoDA-Nets and discuss two of their key properties in sec.~\ref{subsec:align_units}. Concretely, we show that these units linearly transform their inputs with dynamic (input-dependent) weight vectors and, additionally, that they are biased to align these weights with the input during optimisation. 
We then discuss how DAUs can be used for classification (sec.~\ref{subsec:classification}) and how we build performant networks out of multiple layers of convolutional DAUs (sec.~\ref{subsec:coda}). Importantly, the resulting \emph{linear decompositions} of the network outputs are optimised to align with discriminative patterns in the input, making them highly suitable for interpreting the network predictions. 

\begin{figure}[t!]
    \centering
    \hspace{-.25em}
    \begin{subfigure}[b]{0.48\textwidth}
    \includegraphics[width=\textwidth]{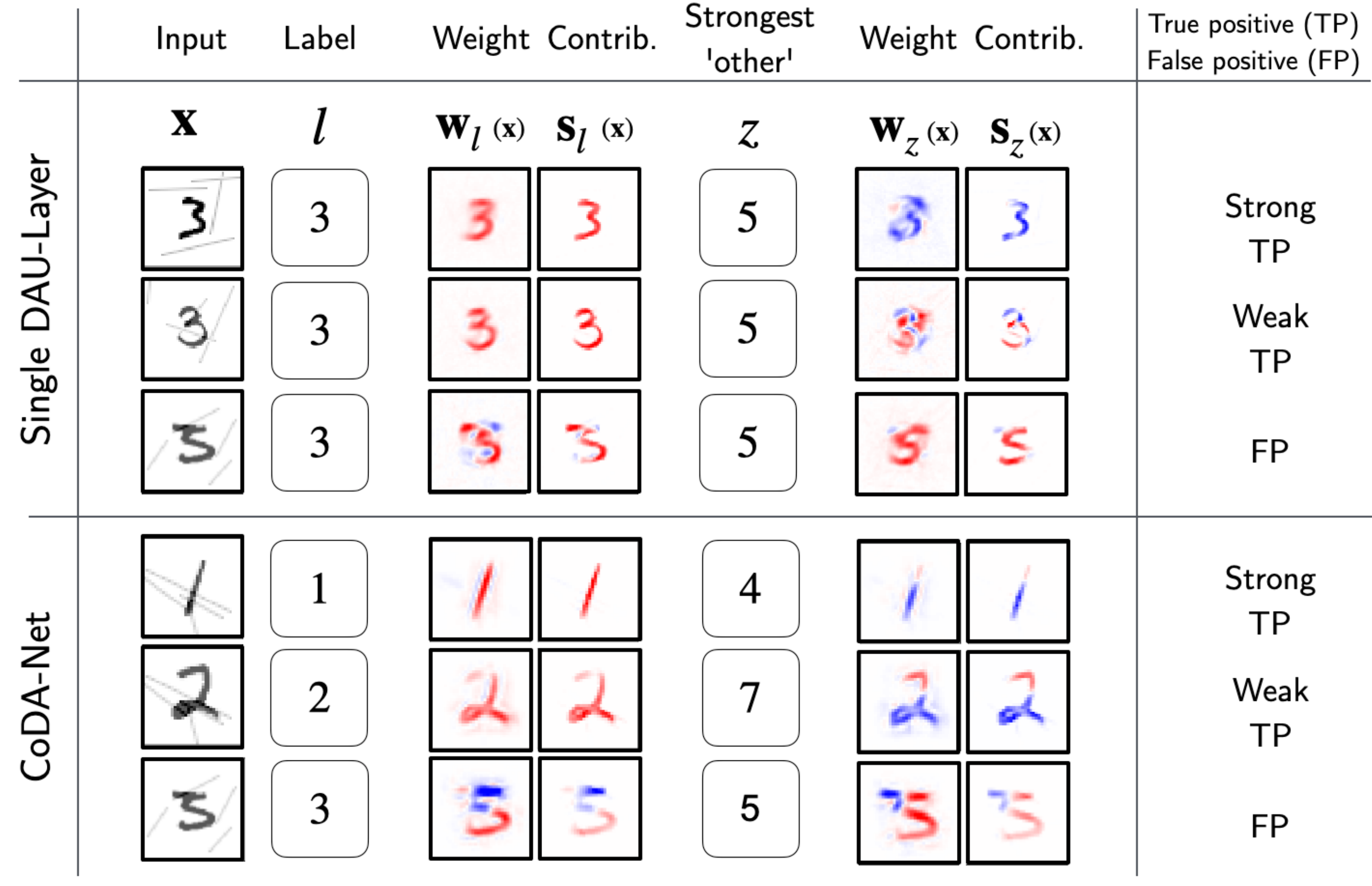}
     \end{subfigure}
    \caption{\small
        For different inputs $\vec x$, we visualise the linear weights and contributions (for the single layer, see eq.~\eqref{eq:contrib_1}, for the CoDA-Net eq.~\eqref{eq:contrib}) for the ground truth label $l$ and the strongest non-label output $z$. 
    As can be seen, the weights align well with the input images.
    The first three rows are based on a single DAU layer, the last three on a 5 layer CoDA-Net. The first two samples (rows) per model are correctly classified and the last one is misclassified. }
    \label{fig:alignment}
\end{figure}

In particular, we structure this section around the following \textbf{three important properties} (\colornum{P1-P3}) of the DAUs:
\\[.25em]
\colornum{P1: Dynamic linearity.} The DAU output $o$ is computed as a dynamic (input-dependent) linear transformation of the input $\vec x$, such that \mbox{$o=\vec w(\vec x)^T\vec x=\sum_jw_j(\vec x)x_j$}. 
Hence, 
$o$ can be decomposed into contributions  
from individual input dimensions, which are given by $w_j(\vec x)x_j$ for dimension $j$.
\\[.5em]
\colornum{P2: Alignment maximisation.} Maximising the average output of a single DAU over a set of inputs $\vec x_i$ 
maximises the alignment between inputs $\vec x_i$ and the weight vectors $\vec w(\vec x_i)$. As the modelling capacity of $\vec w(\vec x)$ is restricted, $\vec w(\vec x)$ will encode the most frequent patterns in the set of inputs $\vec x_i$.
\\[.5em]
\colornum{P3: Inheritance.} When combining multiple DAU layers to form a \mbox{Dynamic} Alignment Network (DA-Net), the properties \colornum{P1} and \colornum{P2} are \emph{inherited}. In particular, DA-Nets are dynamic linear (\colornum{P1}) and maximising the last layer's output induces an output maximisation in the constituent DAUs (\colornum{P2}).
\\[.5em]
These properties increase the interpretability
of a DA-Net, such as a CoDA-Net (sec.~\ref{subsec:coda}) for the following reasons.
First, the output of a DA-Net can be decomposed into contributions from the individual input dimensions, similar to linear models (cf.~Fig.~\ref{fig:teaser}, \colornum{P1} and \colornum{P3}).
Second, we note that optimising a neural network for classification applies a maximisation to the outputs of the last layer for every sample. 
This maximisation aligns the dynamic weight vectors $\vec w(\vec x)$ of the constituent DAUs of the DA-Net with their respective inputs (cf.~Fig.~\ref{fig:alignment}, \colornum{P2} and \colornum{P3}).

 Importantly, the weight vectors will align with the \emph{discriminative} patterns in their inputs when optimised for classification as we show in sec.~\ref{subsec:classification}.
As a result, the model-inherent contribution maps of CoDA-Nets are optimised to align well with \emph{discriminative input patterns} in the input image 
and the interpretability of our models thus forms part of the global optimisation procedure.
\subsection{Dynamic Alignment Units}
\label{subsec:align_units}
We define the Dynamic Alignment Units (DAUs) by
\begin{align}
    \label{eq:au}
    \text{DAU}(\vec x) = g(\mat a \mat b\vec x +\vec b)^T \vec x = \vec w(\vec x)^T\, \vec x\quad \textbf{.}
\end{align}
Here, $\vec x\in\mathbb R^{d}$ is an input vector, $\mat a\in\mathbb R^{d\times r}$ and $\mat b \in \mathbb R^{r\times d}$ are trainable transformation matrices, $\vec b\in\mathbb R^{d}$ a trainable bias vector, and \mbox{$g(\vec u)=\alpha(||\vec u||)\vec u$} is a non-linear function that scales the norm of its input. {In contrast to using a single matrix $\mat m \in\mathbb R^{d\times d}$, using $\mat{ab}$ allows us to control the maximum rank $r$ of the transformation and to reduce the number of parameters}; we will hence refer to $r$ as the rank of a DAU. 
As can be seen by the right-hand side of eq.~\eqref{eq:au}, the DAU linearly transforms the input $\vec x$ (\colornum{P1}). At the same time, given the quadratic form ($\vec x^T\mat B^T\mat A^T\vec x$) and the  rescaling function $\alpha(||\vec u||)$, the output of the DAU is a non-linear function of its input. In this work, we focus our analysis on 
two choices for $g(\vec u)$ in particular\footnote{
In preliminary experiments we observed comparable behaviour over a range of different normalisation functions such as, e.g., L1 normalisation.}, namely rescaling to unit norm ($\text{L2}$) and the squashing function ($\text{SQ}$, see \cite{sabour2017dynamic}):
\begin{align}
    \label{eq:nonlin}
    \text{L2}(\vec u) = \frac{\vec u}{||\vec u||_2} \;\;\text{and}\;\;
    \text{SQ}(\vec u) = \text{L2}(\vec u) \times \frac{||\vec u||^2_2}{1+||\vec u||_2^2}
\end{align}
Under these rescaling functions, the norm of the weight vector is upper-bounded: $||\vec w(\vec x)|| \leq 1$. Therefore, the output of the DAUs is upper-bounded by the norm of the input:
\begin{align}
    \text{DAU}(\vec x) = 
    ||\vec w(\vec x)|| \hspace{.2em} ||\vec x|| \cos(\angle(\vec x, \vec w(\vec x)))\leq ||\vec x||
    \label{eq:bound}
\end{align}
As a corollary, for a given input $\vec x_i$, the DAUs can only achieve this upper bound if $\vec x_i$ is an eigenvector (EV) of the linear transform $\mat{AB}\vec x+ \vec b$. Otherwise, the cosine in eq.~\eqref{eq:bound} will not be maximal\footnote{
Note that $\vec w(\vec x)$ is proportional to $\mat{ab}\vec x + \vec b$. The cosine in eq.~\eqref{eq:bound}, in turn, is maximal if and only if $\vec w(\vec x_i)$ is proportional to $\vec x_i$ and thus, by transitivity, if $\vec x_i$ is proportional to $\mat{ab}\vec x_i + \vec b$. This means that $\vec x_i$ has to be an EV of $\mat{ab}\vec x +\vec b$ to achieve maximal output.}. 
As can be seen in eq.~\eqref{eq:bound}, maximising the average output of a DAU over a set of inputs $\{\vec x_i|\,i=1, ..., n\}$
maximises the alignment between $\vec w(\vec x)$ and $\vec x$ (\colornum{P2}).
In particular, it optimises the parameters of the DAU such that the \emph{most frequent input patterns} are encoded as EVs in the linear transform $\mat{ab}\vec x + \vec b$, similar to an $r$-dimensional PCA decomposition ($r$ the rank of $\mat{ab}$). In fact, as discussed in the supplement, the optimum of the DAU maximisation solves a low-rank matrix approximation~\cite{eckart1936approximation} problem similar to singular value decomposition.
\begin{figure}[t!]
    \centering
    \includegraphics[height=6.5em]{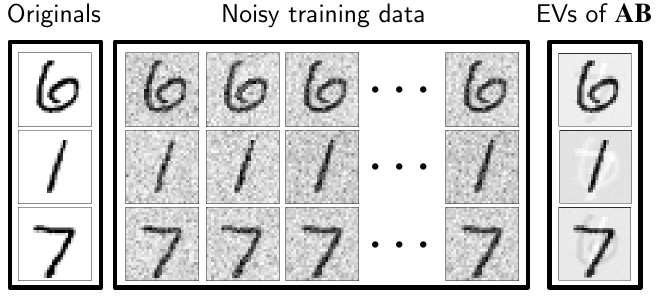}
    \caption{\small Eigenvectors (EVs) of \tmat{AB} after maximising the output of a rank-3 DAU over a set of noisy samples of 3 MNIST digits. Effectively, the DAUs encode the most frequent components in their EVs, similar to a principal component analysis (PCA).
    }
    \label{fig:EVs}
\end{figure}
As an illustration of this property, in Fig.~\ref{fig:EVs} we show the 3 EVs\footnote{Given $r=3$, the EVs maximally span a 3-dimensional subspace.} of matrix $\mat{ab}$ (with rank $r=3$, bias $\vec b=\vec 0$) after optimising a DAU over a set of $n$ noisy samples of 3 specific MNIST~\cite{lecun2010mnist} images; for this, we used $n=3072$ and zero-mean Gaussian noise. As expected, the EVs of \tmat{ab} encode the original, noise-free images, since this on average maximises the alignment (eq.~\eqref{eq:bound}) between the weight vectors $\vec w(\vec x_i)$ and the input samples $\vec x_i$ over the dataset.

\subsection{DAUs for classification}
\label{subsec:classification}
{DAUs can be used directly for classification by applying $k$ DAUs in parallel to obtain an output \mbox{$\hat{\vec y}(\vec x)=\left[\text{DAU}_1(\vec x), ..., \text{DAU}_k(\vec x)\right]$}. 
Note that this is a linear transformation $\hat{\vec y}(\vec x)$$=$$\mat W(\vec x) \vec x$, with each row in $\mat w$$\in$$\mathbb R^{k \times d}$ corresponding to the weight vector $\vec w_j^T$ of a specific DAU $j$.
In particular, consider 
a dataset $\mathcal D = \{(\vec x_i, \vec y_i)|\, \vec x_i\in\mathbb R^d, \vec y_i\in\mathbb R^k\}$ of $k$ classes with `one-hot' encoded labels $\vec y_i$ for the inputs $\vec x_i$.
To optimise the DAUs as classifiers on $\mathcal D$,} we can apply a sigmoid non-linearity to each DAU output and optimise the loss function $\mathcal L = \sum_i\text{BCE}(\sigma(\hat{\vec y}_i), \vec y_i)$, where \text{BCE} denotes the binary cross-entropy and $\sigma$ applies the sigmoid function to each entry in $\hat{\vec y}_i$. Note that for a given sample, \text{BCE} either maximises (DAU for correct class) or minimises (DAU for incorrect classes) the output of each DAU. Hence, this classification loss will still maximise the (signed) cosine between the weight vectors $\vec w(\vec x_i)$ and $\vec x_i$. 

To illustrate this property, in Fig.~\ref{fig:alignment} (top) we show the weights $\vec w(\vec x_i)$ for several samples of the digit `3' after optimising the DAUs for classification on a noisy MNIST dataset; the first two are correctly classified, the last one is misclassified as a `5'. As can be seen, the weights align with the respective input (the weights for different samples are different). However,  different parts of the input are either positively or negatively correlated with a class, which is reflected in the weights: for example, the extended stroke on top of the `3' in the misclassified sample is assigned \emph{negative weight} and, since the background noise is \emph{uncorrelated} with the class labels, it is not represented in the weights. 

In a classification setting, the DAUs {thus} encode \emph{the most frequent discriminative patterns} in the linear transform $\mat{ab}\vec x + \vec b$ such that the dynamic weights $\vec w(\vec x)$ align well with these patterns.
Additionally, since the output for class $j$ is a linear transformation of the input (\colornum{P1}), we can compute the contribution vector $\vec s_j$ containing the per-pixel contributions to this output by the element-wise product ($\odot$)
\begin{align}
\label{eq:contrib_1}
    \vec s_j(\vec x_i) = \vec w_j(\vec x_i)\odot\vec x_i\quad ,
\end{align}
 see Figs.~\ref{fig:teaser} and
\ref{fig:alignment}. 
Such linear decompositions constitute the model-inherent `explanations' which we evaluate in sec.~\ref{sec:results}.
\subsection{Convolutional Dynamic Alignment Networks}
\label{subsec:coda}
The modelling capacity of a single layer of DAUs is limited, similar to a single linear classifier. However, DAUs can be used as the basic building block for deep convolutional neural networks, which yields powerful classifiers. Importantly, in this section we show that such a Convolutional Dynamic Alignment Network (CoDA-Net) inherits the properties (\colornum{P3}) of the DAUs by maintaining both the dynamic linearity (\colornum{P1}) as well as the alignment maximisation (\colornum{P2}). For a convolutional dynamic alignment layer, each filter is modelled by a DAU, similar to dynamic local filtering layers~\cite{jia2016dynamic}. Note that the output of such a layer is also a dynamic linear transformation of the input to that layer, since a convolution is equivalent to a linear layer with certain constraints on the weights, cf.~\cite{convlin}. We include the implementation details in the supplement.
Finally, at the end of this section, we highlight an important difference between output maximisation and optimising for classification with the {BCE} loss. In this context we discuss the effect of \emph{temperature scaling} and present the loss function we optimise in our experiments.

\myparagraph{Dynamic linearity (\colornum{P1}).} In order to see that the linearity is maintained, we note that the successive application of multiple layers of DAUs also results in a dynamic linear mapping. Let $\mat W_l$ denote the linear transformation matrix produced by a layer of DAUs and let $\vec a_{l-1}$ be the input vector to that layer; as mentioned before, each row in the matrix $\mat w_l$ corresponds to the weight vector of a single DAU\footnote{
Note that this also holds for convolutional DAU layers. Specifically, each row in the matrix $\mat w_l$ corresponds to a single DAU applied to exactly one spatial location in the input and the input with spatial dimensions is vectorised to yield $\vec a_{l-1}$. For further details, we kindly refer the reader to~\cite{convlin} and the implementation details in the supplement of this work.}. As such, the output of this layer is given by 
\begin{align}
    \vec a_l = \mat W_l (\vec a_{l-1}) \vec a_{l-1}\quad .
\end{align}
In a network of DAUs, the successive linear transformations can thus be collapsed. In particular, \emph{for any pair of activation vectors} $\vec{a}_{l_1}$ and $\vec{a}_{l_2}$ with ${l_1}<{l_2}$, the vector $\vec{a}_{l_2}$ can 
    be expressed as a linear transformation of $\vec{a}_{l_1}$:
\begin{align}
\label{eq:collapse}
    \vec{a}_{l_2} &= \mat{W}_{{l_1}\rightarrow {l_2}} \left(\vec{a}_{l_1}\right)\vec{a}_{l_1} \quad 
        \\{with} \quad \mat{W}_{{l_1}\rightarrow {l_2}}\left(\vec{a}_{l_1}\right) &= \textstyle\prod_{k={l_1}+1}^{l_2} \mat{W}_k \left(\vec{a}_{k-1}\right)\quad \text{.}
\end{align}
For example, the matrix $\mat W_{0\rightarrow L}(\vec{a}_0 = \vec{x}) = \mat W(\vec{x})$ models the linear transformation from the input to the output space, see Fig.~\ref{fig:teaser}.
Since this linearity holds between any two layers, the $j$-th entry of any activation vector $\vec a_l$ in the network can be decomposed into input contributions via:
    \begin{align}
    \label{eq:contrib}
        \vec{s}_{j}^l(\vec x_i) = \left[\mat W_{0\rightarrow l} (\vec{x}_i)\right]_j^T \odot \vec x_i\quad \text{,}
    \end{align}
    with $[\mat W]_j$ the $j$-th row in the matrix.

\myparagraph{Alignment maximisation (\colornum{P2}).}
Note that the output of a CoDA-Net is bounded independent of the network parameters: since each DAU operation can---independent of its parameters---at most reproduce the norm of its input (eq.~\eqref{eq:bound}), the linear concatenation of these operations necessarily also has an upper bound which does not depend on the parameters.
Therefore, in order to achieve maximal outputs on average (e.g., the class logit over the subset of images of that class), all DAUs in the network need to produce weights $\vec w (\vec a_l)$ that align well with the class features. In other words, the weights will align with discriminative patterns in the input.
For example, in Fig.~\ref{fig:alignment} (bottom), we visualise the `global matrices' $\mat W_{0\rightarrow L}$ and the corresponding contributions (eq.~\eqref{eq:contrib}) for a $L=5$ layer CoDA-Net. As before, the weights align with discriminative patterns in the input and do not encode the uninformative noise.

\myparagraph[0]{Temperature scaling and loss function.} 
\begin{figure}[t]
    \centering
    \includegraphics[width=.45\textwidth]{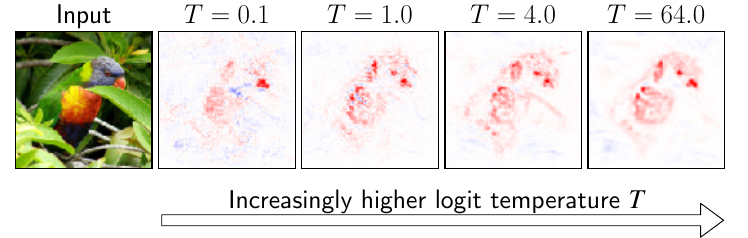}
    \caption{\small By lowering the upper bound (cf.~eq.~\eqref{eq:bound}), the correlation maximisation in the DAUs can be emphasised.
    We show contribution maps for a model trained with different temperatures.
    }
    \label{fig:scaling}
\end{figure}
So far we have assumed that minimising the {BCE} loss for a given sample is equivalent to applying a maximisation or minimisation loss to the individual outputs of a CoDA-Net. While this is in principle correct, {BCE} introduces an additional, non-negligible effect: \emph{saturation}. Specifically, it is possible for a CoDA-Net to achieve a low {BCE} loss without the need to produce well-aligned weight vectors. As soon as the classification accuracy is high and the outputs of the networks are large, the gradient---and therefore the \emph{alignment pressure}---will vanish. This effect can, however, easily be mitigated:
 as discussed in the previous paragraph, the output of a CoDA-Net is upper-bounded \textit{independent of the network parameters}, since each individual DAU in the network is upper-bounded. 
By scaling the network output with a temperature parameter $T$ such that 
    $\hat{\vec y} (\vec x) = T^{-1} \mat W_{0\rightarrow L}(\vec x)\,\vec x$, 
we can explicitly decrease this upper bound and thereby increase the \emph{alignment pressure} in the DAUs by avoiding the early saturation due to {BCE}.
In particular, the lower the upper bound is, the stronger the induced DAU output maximisation should be, since the network needs to accumulate more signal to obtain large class logits (and thus a negligible gradient). This is indeed what we observe both qualitatively, cf.~Fig.~\ref{fig:scaling}, and quantitatively, cf.~Fig.~\ref{fig:localisation} (right column).
Alternatively, the representation of the network's computation as a linear mapping allows to directly regularise what properties these linear mappings should fulfill. For example, we show in the supplement that by regularising the absolute values of the matrix $\mat W_{0\rightarrow L}$, we can induce sparsity in the signal alignments, which can lead to sharper heatmaps.
The overall loss for an input $\vec x_i$ and the target vector $\vec y_i$ is thus computed as 
    \begin{align}
        \label{eq:loss}
        \mathcal{L}(\vec x_i, \vec y_i) &= 
        \text{BCE}(\sigma(T^{-1} \mat W_{0\rightarrow L}(\vec x_i)\,\vec{x}_i + {\vec{b}}_0)\,,\, \vec{y}_i) \\&+ 
        \lambda \langle | \mat W_{0\rightarrow L}(\vec x_i) |\rangle\quad \text{.}
    \end{align}
    Here, $\lambda$ is the strength of the regularisation, $\sigma$ applies the sigmoid activation to each vector entry,
    ${\vec{b}}_0$ is a fixed bias term, and $\langle|\mat W_{0\rightarrow L}(\vec x_i)|\rangle$ refers to the mean over the absolute values of 
        all entries in the matrix $\mat W_{0\rightarrow L}(\vec x_i)$.
\subsection{Implementation details}
\label{subsec:details}
\myparagraph[-.25]{Shared matrix \tmat b.} In our experiments, we opted to share the matrix $\mat b\in \mathbb R^{r\times d}$ between all DAUs in a given layer. This increases parameter efficiency by having the DAUs share a common $r$-dimensional subspace and still fixes the maximal rank of each DAU to the chosen value of $r$. 

\myparagraph[-.25]{Input encoding.} 
In sec.~\ref{subsec:align_units}, we showed that the norm-weighted cosine similarity between the dynamic weights and the layer inputs is optimised and the output of a DAU is at most the norm of its input. This favours pixels with large RGB values, since these have a larger norm and can thus produce larger outputs in the maximisation task. To mitigate this bias, we add the negative image as three additional color channels and thus encode each pixel in the input 
as 
\mbox{[$r$, $g$, $b$, $1-r$, $1-g$, $1-b$]}, with $r, g, b\in [0, 1]$.

\section{Results}
\label{sec:results}
\begin{figure*}[t]
\centering

\fcolorbox[rgb]{0.43921569, 0.50196078, 0.56470588}{0.9176470588235294, 0.9176470588235294, 0.9490196078431372}{
\begin{subfigure}[c]{.95\textwidth}
\includegraphics[width=1\textwidth]{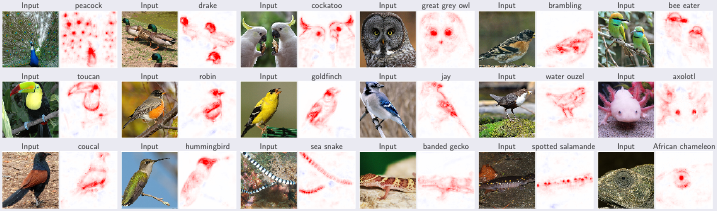}
\end{subfigure}
}
            \caption{\small Model-inherent contribution maps for the most confident predictions for 18 different classes, sorted by confidence (high to low). We show positive (negative) contributions (eq.~\eqref{eq:contrib}) per spatial location for the ground truth class logit in red (blue).}
            \label{fig:quality}
\end{figure*}
In sec.~\ref{subsec:accuracy}, we describe the experimental setup, assess the classification performance of the CoDA-Nets and discuss their efficiency.
Further, in sec.~\ref{subsec:intp_results} we evaluate the model-inherent contribution maps derived from $\mat w_{0\rightarrow L}$ (cf.~eq.~\eqref{eq:contrib}) and compare them both \emph{qualitatively} (Fig.~\ref{fig:quality}) as well as \emph{quantitatively} (Fig.~\ref{fig:localisation}) to other attribution methods.

\subsection{Setup and model performance}
\label{subsec:accuracy}
\begin{table}
    \centering
    {\setlength{\tabcolsep}{0.25em}\setlength\extrarowheight{-1pt}
    \begin{tabular}{>{\centering\arraybackslash} >{\centering\arraybackslash}p{2.85cm} | >{\centering\arraybackslash}p{.75cm} >{\centering\arraybackslash}p{.1cm}  >{\centering\arraybackslash}p{2.85cm} | >{\centering\arraybackslash}p{.75cm} }
    \footnotesize \textbf{Model} & \footnotesize \textbf{C10}
    && \footnotesize \textbf{Model} & \footnotesize \textbf{T-IM}\\[.2em]
    \cline{1-2}
    \cline{4-5}
    & \footnotesize&& & \footnotesize\\[-.8em]
    \footnotesize SENNs~\cite{melis2018towards} & \footnotesize 78.5\% &
    & \footnotesize ResNet-34~\cite{resnet_tiny} & \footnotesize 52.0\%
    \\
    \footnotesize VGG-19~\cite{maxgain} & \footnotesize 91.5\%&
    & \footnotesize VGG~16~\cite{vgg_tiny} & \footnotesize 52.2\%
    \\
    \footnotesize DE-CapsNet~\cite{jia2020capsnet} & \footnotesize 93.0\%&
    & \footnotesize VGG~16  + aug \cite{vgg_tiny} & \footnotesize 56.4\%
    \\
    \footnotesize ResNet-56~\cite{he2016deep} & \footnotesize 93.6\%&
    & \footnotesize IRRCNN~\cite{irrcnn_alom} & \footnotesize 52.2\%
    \\
    \footnotesize WRN-28-2~\cite{he2016deep} & \footnotesize 94.9\%&
    & \footnotesize ResNet-110~\cite{rn110_tiny} & \footnotesize 56.6\%
    \\
    \footnotesize WRN-28-2 + aug~\cite{cubuk2019randaugment} & \footnotesize 95.8\%&
    & \footnotesize WRN-40-20~\cite{hendrycks_tiny} & \footnotesize 63.8\%
    \\[.2em]
    \cline{1-2}\cline{4-5}
    &&&&\\[-.75em]
    \footnotesize S-CoDA-SQ ($\lambda$)& \footnotesize 93.8\%&
    & \footnotesize XL-CoDA-SQ ($T$)& \footnotesize 54.4\%
    \\
    \footnotesize S-CoDA-L2 ($\lambda$)& \footnotesize 92.6\%&
    & \footnotesize XL-CoDA-SQ + aug ($T$)& \footnotesize 58.4\%
    \\
    \footnotesize S-CoDA-SQ ($T$)& \footnotesize 93.2\%&
    &&\\
    \footnotesize S-CoDA-L2 ($T$)& \footnotesize 93.0\%
    &&
    \\
    \footnotesize M-CoDA-SQ + aug ($\lambda$)& \footnotesize 96.5\%&
    & &
    \end{tabular}}
    \caption{\small CIFAR-10 (\textbf{C10}) and TinyImagenet (\textbf{T-IM}) 
    classification accuracies. Results taken from specified references. The prefix of the CoDAs indicates model size, the suffix the non-linearity used (eq.~\eqref{eq:nonlin}). With ($\lambda$) and ($T$) we denote if models were trained with regularisation or increased temperature $T$, see eq.~\eqref{eq:loss}.}
    \label{tbl:result_table}
\end{table}

\myparagraph[-.25]{Datasets.} We evaluate and compare the accuracies of the CoDA-Nets to other work on the CIFAR-10~\cite{krizhevsky2009cifar10} and the TinyImagenet~\cite{tinyimagenet} datasets. We use the same datasets for the quantitative evaluations of the model-inherent contribution maps. Additionally, we qualitatively show high-resolution examples from a CoDA-Net trained on the first 100 classes of the Imagenet dataset. 

\myparagraph[-.25]{Models.} 
We evaluate models of four different sizes
denoted by (S/M/L/XL)-CoDA on CIFAR-10 (S and M), Imagenet-100 (L), and TinyImagenet (XL); these models have 8M (S), 28M (M), 48M (L), and 62M (XL) parameters respectively; see the supplement for an evaluation of the impact of model size on accuracy.
All models feature 9 convolutional DAU layers and a final sum-pooling layer, and mainly vary in the number of features, the rank $r$ of the DAUs,  and the convolutional strides for reducing the spatial dimensions. 
No additional methods such as residual connections, dropout, or batch normalisation are used. This 9-layer architecture was initially optimised for the CIFAR-10 dataset and subsequently adapted to the TinyImagenet and Imagenet-100 datasets. 
Further, we investigate the effect that the temperature $T$, the regularisation $\lambda$, and the non-linearities (\text{L2}, \text{SQ}, see eq.~\eqref{eq:nonlin}) have on the CoDA-Nets. Given the computational cost of the regularisation (two additional passes to extract and regularise $\mat w_{0\rightarrow L}$), evaluate the regularisation on models trained on CIFAR-10. Lastly, models marked with $T$ ($\lambda$) in Table \ref{tbl:result_table} were trained with $\lambda$$=$$0$ ($T$$=$$64$, equiv.~to `average pooling'). Details on architectures and training procedure are included in the supplement.

\myparagraph{Classification performance.} In Table \ref{tbl:result_table} we compare the performances of our CoDA-Nets to several other published results. Note that the referenced numbers are meant to be used as a gauge for assessing the CoDA-Net performance and do not exhaustively represent the state of the art. In particular, we would like to highlight that the CoDA-Net performance is on par to models of the VGG~\cite{vgg} and ResNet~\cite{he2016deep} model families on both datasets. Moreover, under the same data augmentation (RandAugment~\cite{cubuk2019randaugment}), it achieves similar results as the WideResNet-28-2~\cite{zagoruyko2016wide} on CIFAR-10.
Additionally, we list the reported results of the SENNs~\cite{melis2018towards} and the DE-CapsNet~\cite{jia2020capsnet} architectures for CIFAR-10. Similar to our CoDA-Nets, the SENNs were designed to improve network interpretability and are also based on the idea of explicitly modelling the output as a dynamic linear transformation of the input. On the other hand, the CoDA-Nets share similarities to capsule networks, which we discuss in the supplement; to the best of our knowledge, the \mbox{DE-CapsNet} currently achieves the state of the art in the field of capsule networks on CIFAR-10. 
Overall, we observed that the CoDA-Nets deliver competitive performances that are fairly robust to the non-linearity (\text{L2}, \text{SQ}), the temperature ($T$), and the regularisation strength ($\lambda$). We note that on average SQ performed better than L2, which we ascribe to the fact that SQ avoids up-scaling vectors with low norm ($||\vec v||<1$, see eq.~\eqref{eq:nonlin}).

\myparagraph[-.3]{Efficiency considerations.} 
The CoDa-Nets achieve good accuracies on the presented datasets, exhibit training behaviour that is robust over a wide range of hyperparameters, and are as fast as a typical ResNet at inference time.
However, under the current formulation and without highly optimised GPU implementations for the DAUs, training times are significantly longer for the CoDA-Nets.
While we are currently working on an improved and optimised version of CoDA-Nets, we were not yet able to generate results for the full ImageNet dataset.
On the 100 classes subset, however, the evaluated L-CoDA-SQ network achieved competitive performance (76.5\% accuracy, for details see supplement) and offers highly detailed explanations for its predictions, as we show in Figs.~\ref{fig:quality} and \ref{fig:comparison}.

\subsection{Interpretability of CoDA-Nets}
\label{subsec:intp_results}
\begin{figure*}
%
    \centering
    \begin{subfigure}[c]{0.32\textwidth}
    \centering
    \includegraphics[width=\textwidth, trim=0 1em 0 1em, clip]{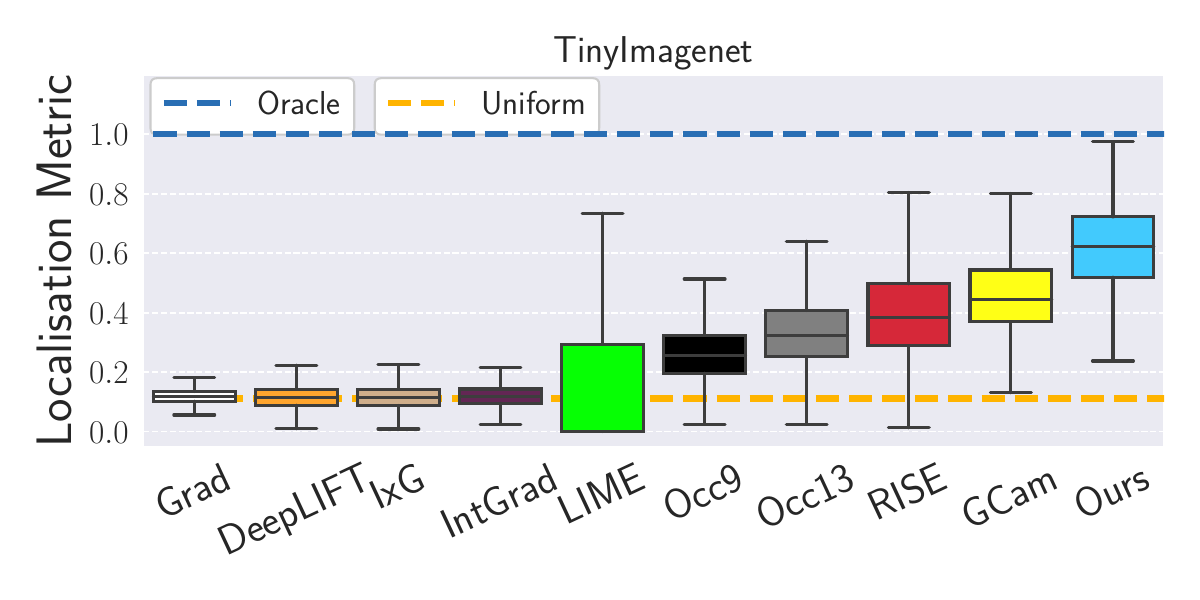}
    \end{subfigure}
    \begin{subfigure}[c]{0.32\textwidth}
    \centering
    \includegraphics[width=\textwidth, trim=0 1em 0 1em, clip]{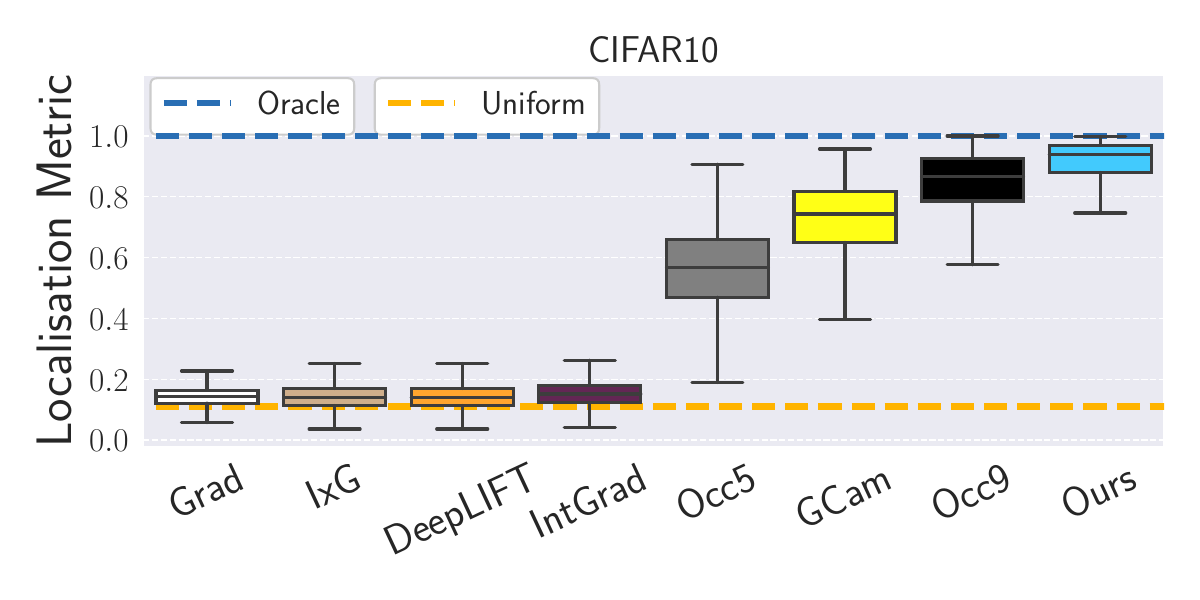}
    \end{subfigure}
    \begin{subfigure}[c]{0.32\textwidth}
    \centering
    \includegraphics[width=\textwidth, trim=0 1em 0 1em, clip]{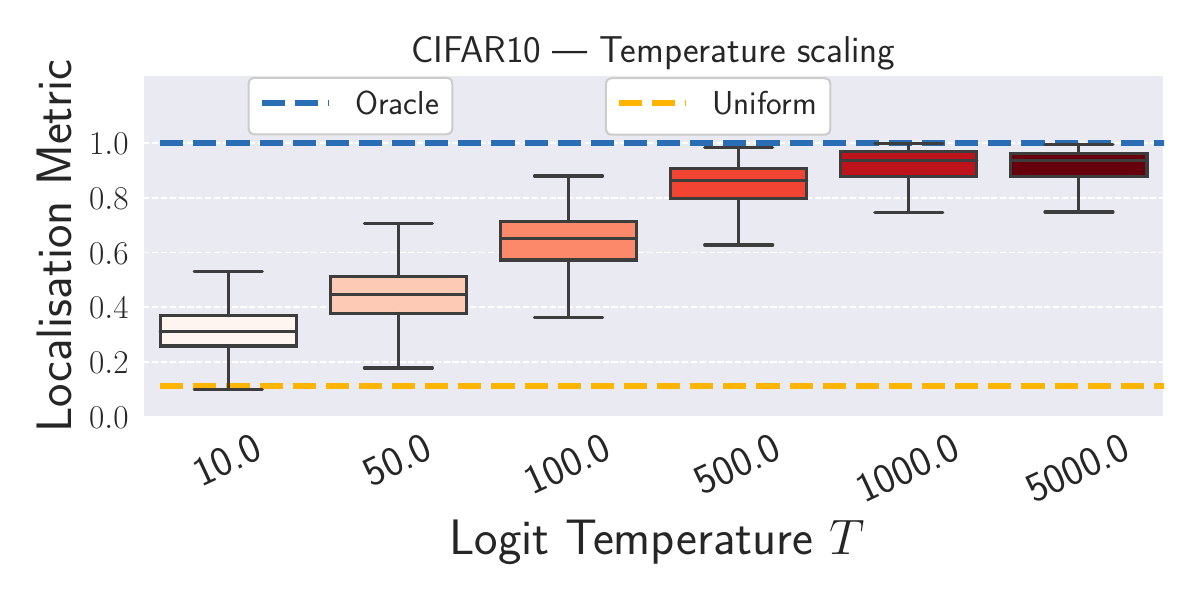}
    \end{subfigure}
    \begin{subfigure}[c]{0.320\textwidth}
    \centering
    \includegraphics[width=\textwidth, trim=0 1em 0 1em, clip]{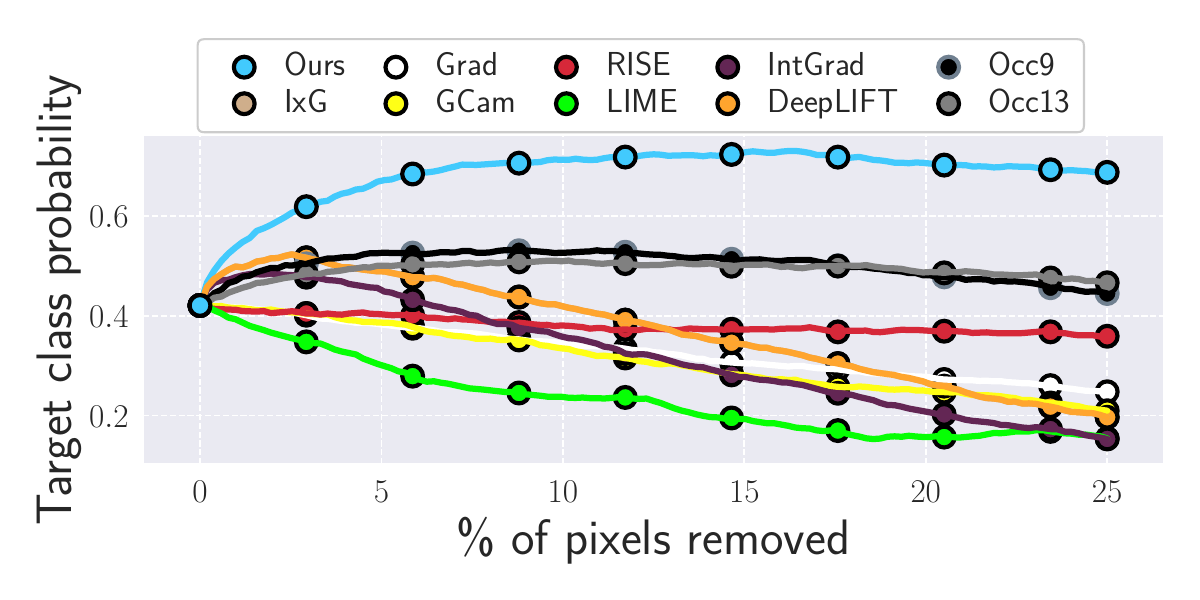}
    \end{subfigure}
    \begin{subfigure}[c]{0.320\textwidth}
    \centering
    \includegraphics[width=\textwidth, trim=0 1em 0 1em, clip]{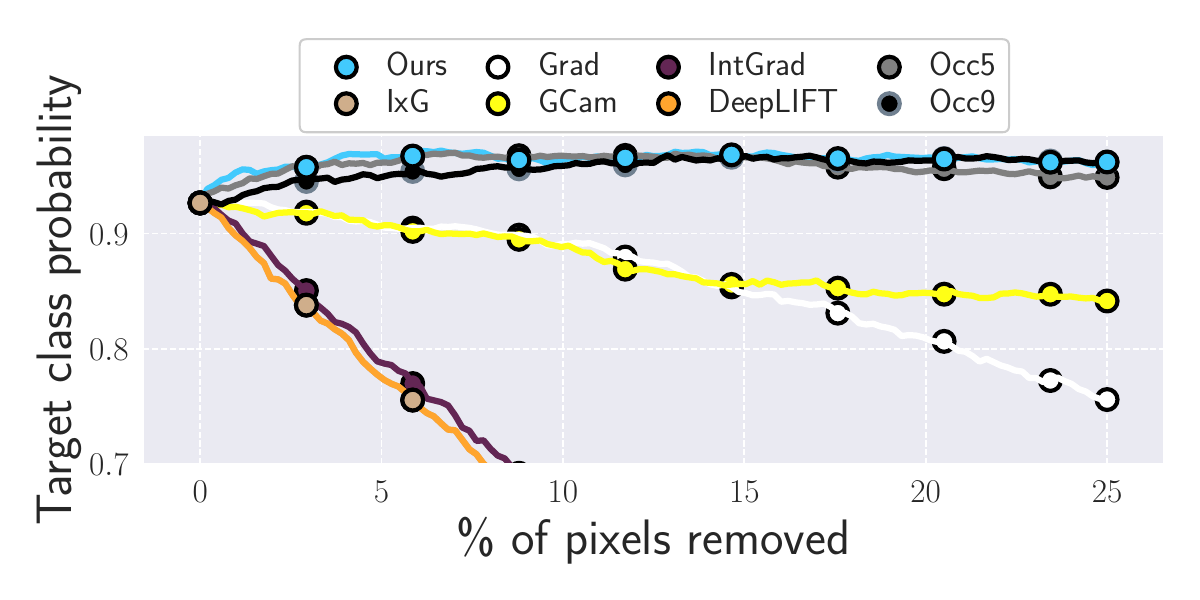}
    \end{subfigure}
    \begin{subfigure}[c]{0.320\textwidth}
    \centering
    \includegraphics[width=\textwidth, trim=0 1em 0 1em, clip]{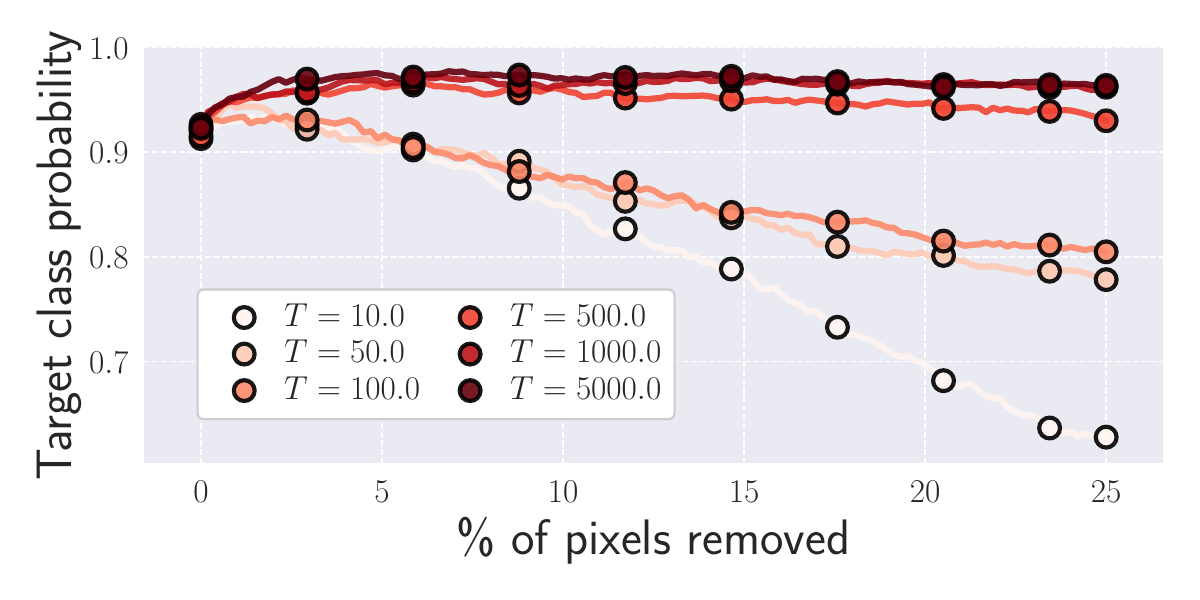}
    \end{subfigure}
    \caption{\small 
    \textbf{Top row:} Results for the localisation metric, see eq.~\eqref{eq:loc_metric}.
    \textbf{Bottom row:} Pixel removal metric. In particular, we plot the mean target class probability  after removing the $x\%$ of the \emph{least important} pixels.
    We show the results of a CoDA-Net trained on TinyImagenet \textbf{(left column)}, as well as on CIFAR-10 \textbf{(center column)}.
    Additionally, we show the effect of the temperature parameter on the interpretability of the CoDA-Nets \textbf{(right column)}:
    as expected, a higher temperature leads to higher interpretability (sec.~\ref{subsec:details}).
    }
    \label{fig:localisation}
\end{figure*}
In the following, we evaluate the model-inherent contribution maps and compare them to other commonly used me\-thods for importance attribution.
The evaluations are based on the XL-CoDA-SQ (T=6400) for TinyImagenet and the S-CoDA-SQ (T=1000) for CIFAR-10, see~Table~\ref{tbl:result_table} for the respective accuracies.
Further, we evaluate the effect of training the same CIFAR-10 architecture with different temperatures $T$; as discussed in sec.~\ref{subsec:coda}, we expect the interpretability to \emph{increase} along with $T$, since for larger $T$ a stronger alignment is required in order for the models to obtain large class logits. Evaluations of models trained with L1-regularisation of the matrices $\mat m_{0\rightarrow L}$ (eq.~\eqref{eq:loss}) and of models with the L2 non-linearity (eq.~\eqref{eq:nonlin}) are included in the supplement. The respective results are similar to those presented here. Before turning to the results, however, in the following we will first present the attribution methods used for comparison and discuss the evaluation metrics employed for quantifying their interpretability.

\myparagraph{Attribution methods.} We compare the model-inherent contribution maps (cf.~eq.~\eqref{eq:contrib}) to other common approaches for importance attribution. 
In particular, we evaluate against several perturbation based methods such as RISE~\cite{petsiuk2018rise}, LIME~\cite{lime}, and several occlusion attributions~\cite{zeiler2014visualizing} (Occ-K, with K the size of the occlusion patch). Additionally, we evaluate against common gradient-based methods. These include the gradient of the class logits with respect to the input image~\cite{baehrens2010explain} (Grad), `Input$\times$Gradient' (IxG, cf.~\cite{adebayo2018sanity}), GradCam~\cite{selvaraju2017grad} (GCam), Integrated Gradients~\cite{sundararajan2017axiomatic} (IntG), and DeepLIFT~\cite{shrikumar2017deeplift}. As a baseline, we also evaluated these methods on a pre-trained ResNet-56~\cite{he2016deep} on CIFAR-10, for which we show the results in the supplement.

\begin{figure}
    \centering
    \fcolorbox[rgb]{0.43921569, 0.50196078, 0.56470588}{0.9176470588235294, 0.9176470588235294, 0.9490196078431372}
    {
    \centering
    \includegraphics[width=.45\textwidth, trim=2em 1em 1em 0em]{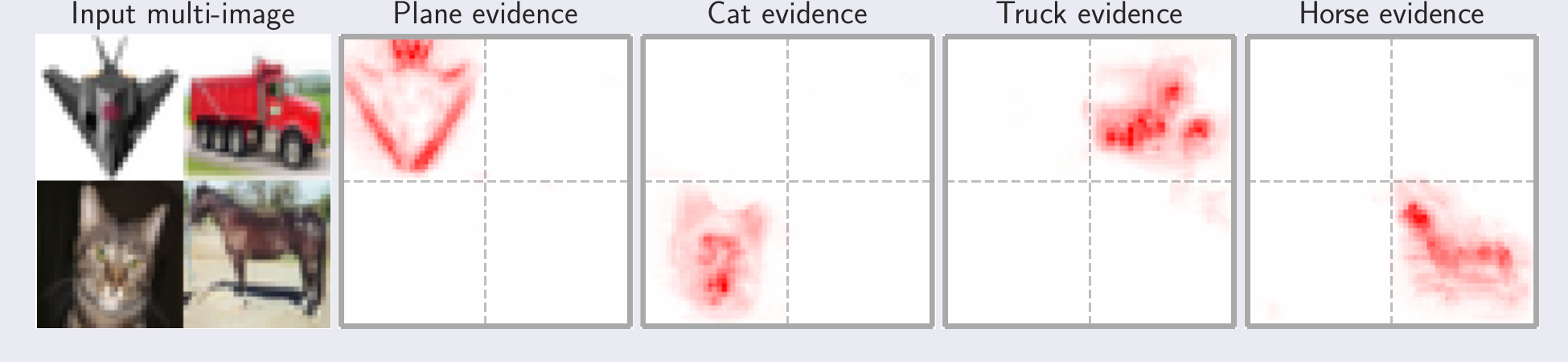}
    }
    \caption{\small A multi-image on the CIFAR-10 dataset. The CoDA-Net contribution maps highlight the individual class-images well.}
    \label{fig:multi_image}
\end{figure}

\myparagraph{Evaluation metrics.}
Our quantitative evaluation of the attribution maps is based on the following two methods:
we
    \mbox{\colornum{(1)} evaluate} a localisation metric by adapting the pointing game~\cite{zhang2018top} to the CIFAR-10 and TinyImagenet datasets, and 
    \mbox{\colornum{(2)} analyse} the model behaviour under the pixel removal strategy employed in~\cite{srinivas2019full}.
For~\colornum{(1)}, 
we evaluate the attribution methods on a grid of $ n\times n$ with $n=3$ images sampled from the corresponding datasets; in every grid of images, each class may occur at most once. 
For a visualisation with $n=2$, see
Fig.~\ref{fig:multi_image}.
\begin{figure}
\centering
\fcolorbox[rgb]{0.43921569, 0.50196078, 0.56470588}{0.9176470588235294, 0.9176470588235294, 0.9490196078431372}{
    \begin{subfigure}[c]{.45\textwidth}
    \centering
    \includegraphics[width=\textwidth]{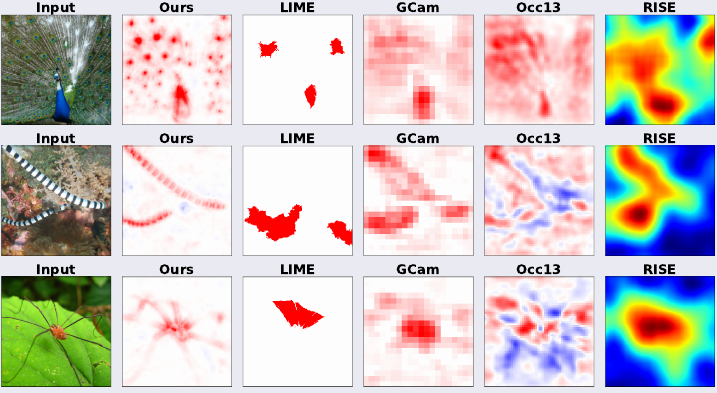}
            \end{subfigure}}
            \caption{\small Comparison to the strongest post-hoc methods. While the regions of importance roughly coincide, the inherent contribution maps of the CoDA-Nets offer the most detail. Note that to improve the RISE visualisation, we chose its default colormap~\cite{petsiuk2018rise}; the most (least) important values are still shown in red (blue).}
            \label{fig:comparison}
\end{figure}
For each occurring class, we can measure how much positive importance an attribution method assigns to the respective class image.
Let $\mathcal{I}_c$ be the image for class $c$, then the score $s_c$ for this class is calculated as 
\begin{align}
\label{eq:loc_metric}
            \textstyle s_c = \frac{1}{Z}\sum_{p_c \in \mathcal{I}_c} p_c \quad \text{with}\quad 
            Z = {\sum_k\sum_{p_c \in \mathcal{I}_k} p_c}\quad ,
        \end{align}
        with $p_c$ the positive attribution for class $c$ assigned to the spatial location $p$.
        This metric has the same clear oracle score  $s_c=1$ for all attribution methods (all positive attributions located in the correct grid image)
        and a clear score for completely random attributions $s_c=1/n^2$ (the positive attributions are uniformly distributed over the different grid images).
        Since this metric depends on the classification accuracy of the models, we sample the first $500$ (CIFAR-10) or $250$ (TinyImagenet) images according to their class score for the ground-truth class\footnote{
        We can only expect an attribution to specifically highlight a class image if this image can be correctly classified on its own. If all grid images have similarly low attributions, the localisation score will be random.
        }; note that since all attributions are evaluated for the same model on the same set of images, this does not favour any particular attribution method.\\
For~\colornum{(2)}, we show how the model's class score behaves under the removal of an increasing amount of \emph{least important} pixels,
    where the importance is obtained via the respective attribution method. Since the first pixels to be removed are typically assigned negative or relatively little importance, we expect the model to initially increase its confidence (removing pixels with \emph{negative} impact) or maintain a similar level of confidence (removing pixels with \emph{low} impact) if the evaluated attribution method produces an accurate ranking of the pixel importance values. 
Conversely, if we were to remove the \emph{most important} pixels first, we would expect the model confidence to quickly decrease. However, as noted by \cite{srinivas2019full}, removing the most important pixels first introduces artifacts in the most important regions of the image and is therefore potentially more unstable than removing the least important pixels first.
Nevertheless, the model-inherent contribution maps perform well in this setting, too, as we show in the supplement.
Lastly, in the supplement we qualitatively show that they pass the `sanity check' of \cite{adebayo2018sanity}.

\myparagraph[0]{Quantitative results.}
In Fig.~\ref{fig:localisation}, we compare the contribution maps of the CoDA-Nets to other attributions under the evaluation metrics discussed above.
It can be seen that the CoDA-Nets~\colornum{(1)}
    perform well under the localisation metric given by eq.~\eqref{eq:loc_metric} and outperform all 
    the other attribution methods evaluated on the same model, both for TinyImagenet (top row, left) and CIFAR-10 (top row, center); note that we excluded RISE and LIME on CIFAR-10, since the default parameters do not seem to transfer well to this low-resolution dataset. 
Moreover, \colornum{(2)} 
    the CoDA-Nets perform well in the pixel-removal setting: 
    the \emph{least salient} locations according to the model-inherent contributions indeed seem to be among the least relevant for the given class score on both datasets, see Fig.~\ref{fig:localisation} (bottom row, left and center). 
Further, in Fig.~\ref{fig:localisation} (right column),  we show the effect of temperature scaling on the interpretability of CoDA-Nets trained on CIFAR-10. The results indicate that the alignment maximisation is indeed crucial for interpretability and constitutes an important difference of the CoDA-Nets to other dynamic linear networks such as piece-wise linear networks (ReLU-based networks). In particular, by structurally requiring a strong alignment for confident classifications, the interpretability of the CoDA-Nets forms part of the optimisation objective.
Increasing the temperature increases the alignment and thereby the interpretability of the CoDA-Nets. While we observe a downward trend in classification accuracy when increasing $T$, the best model at $T=10$ only slightly improved the accuracy compared to $T=1000$ ($93.2\%\rightarrow 93.6\%$); for more details, see supplement.

In summary, the results show that by combining dynamic linearity with a structural bias towards an alignment with discriminative patterns, we obtain models which inherently provide an interpretable linear decomposition of their predictions. 
Further, given that we better understand the relationship between the intermediate computations and the optimisation of the final output in the CoDA-Nets, we can emphasise model interpretability in a principled way by increasing the `alignment pressure' via \emph{temperature scaling}.
    
\myparagraph[0]{Qualitative results.} In Fig.~\ref{fig:quality}, we visualise spatial contribution maps of the L-CoDA-SQ model (trained on Imagenet-100) for some of its most confident predictions. Note that these contribution maps are linear decompositions of the output and the sum over these maps yields the respective class logit. In Fig.~\ref{fig:comparison}, we additionally present a visual comparison to the best-performing post-hoc attribution methods; note that RISE cannot be displayed well under the same color coding and we thus use its default visualisation. We observe that the different methods are not inconsistent with each other and roughly highlight similar regions. However, the inherent contribution maps are of much higher detail and compared to the perturbation-based methods do not require multiple model evaluations. Much more importantly, however, all the other methods are attempts at approximating the model behaviour \emph{post-hoc}, while the CoDA-Net contribution maps in Fig.~\ref{fig:quality} are derived from the model-inherent linear mapping that is used to compute the model output.
\section{Discussion and conclusion}
\label{sec:discussion}
In this work, we presented a new family of neural networks, the CoDA-Nets, 
and show that they are performant classifiers with a high degree of interpretability.
For this, we first introduced the Dynamic Alignment Units, which model their output as a dynamic linear transformation of their input and have a structural bias towards alignment maximisation.
    Using the DAUs to model filters in a convolutional network, we obtain the Convolutional Dynamic Alignment Networks (CoDA-Nets).
The successive linear mappings by means of the DAUs within the network make it possible to linearly decompose the output into contributions from individual input dimensions. 
In order to assess the quality of these contribution maps,
    see eq.~\eqref{eq:contrib}, we compare against other attribution methods.
    We find that the CoDA-Net contribution maps consistently perform well under commonly used quantitative metrics.
    Beyond their \emph{interpretability},
        the CoDA-Nets constitute performant classifiers: their accuracy on CIFAR-10 and the TinyImagenet dataset are on par to the commonly employed VGG and ResNet models.

{\small
\justifying
\bibliographystyle{ieee_fullname}
\bibliography{main_bib}
\justifying
}
\clearpage
{
\numberwithin{equation}{section}
\numberwithin{figure}{section}
\numberwithin{table}{section}
\renewcommand{\thefigure}{\thesection\arabic{figure}}
\renewcommand{\thetable}{\thesection\arabic{table}}

\appendix
\onecolumn 
\flushleft

\begin{center}
{
\huge\bf \vspace{1em}Supplementary Material\\[2em]}
\end{center}

\newcommand{\additem}[2]{%
\item[\textbf{(\ref{#1})}] 
    \textbf{#2} \dotfill\makebox{\textbf{\pageref{#1}}}
}

\newcommand{\addsubitem}[2]{%
    \\[.5em]\indent\hspace{1em}
    \textbf{(\ref{#1})}
    #2 \dotfill\makebox{\textbf{\pageref{#1}}}
}

\newcommand{\adddescription}[1]{\newline
\begin{adjustwidth}{1cm}{1cm}
#1
\end{adjustwidth}
}
{\vspace{2em}\bf\Large Table of Contents\\[2em]}

In this supplement to our work on Convolutional Dynamic Alignment Networks (CoDA-Nets), we provide:
\begin{enumerate}[label={({\arabic*})}, topsep=1em, itemsep=.5em]
    \additem{sec:additional_figures}{ 
    Additional qualitative results} 
    \adddescription{In this section, we show additional \emph{qualitative} results on the Imagenet subset as well as additional comparisons between the model-inherent contribution maps and other methods for importance attribution. Further, we show the effect of regularising the linear mappings on the contribution maps for models trained on the CIFAR-10 dataset. Lastly, we show the results of the `sanity check' by Adebayo et al.~\citesupp{adebayo2018sanity} as well as the contribution maps of a piece-wise linear model (ResNet-56).}
    \additem{sec:additional_quantitative}{ 
    Additional quantitative results} 
    \adddescription{In this section, we show additional \emph{quantitative} results. In particular, we show the accuracies of the temperature-regularised models and of models of different sizes (DAU rank ablation). Further, we show interpretability results for models trained with the L2 non-linearity, with explicit regularisation of the linear mapping $\mat w_{0\rightarrow L}$, and for a pre-trained ResNet-56 for comparison. Moreover, we show results for the \emph{pixel removal metric} when removing the most important pixels first.}
    \additem{sec:training}{ 
    Implementation details} 
    \adddescription{In this section, we present architecture and training details for our experiments and describe in detail how the convolutional Dynamic Alignment Units are implemented. Further, we discuss the results of ResNets on the Imagenet subset under the exact same training scheme for comparison.}
    \additem{sec:low_rank_matrices}{
    Relation to low-rank matrix approximations
    }
    \adddescription{In this section, we discuss the relationship between the Dynamic Alignment Units and the problem of low-rank matrix approximation.
    }
    \additem{sec:capsule_comparison}{
    Relation to capsule networks}
    \adddescription{In this section, we discuss the relationship between the Dynamic Alignment Units and capsules~\citesupp{sabour2017dynamic}. In particular, we rewrite the standard capsule formulation, which allows us to compare them more easily to our work. Under this new formulation, it becomes clear that the two approaches share similarities, but also that there exist important differences.
    }
    
\end{enumerate}
}
\clearpage
\justify
\section{Additional qualitative results}
\label{sec:additional_figures}

\begin{figure*}[t]
    \centering
    \fcolorbox[rgb]{0.9176470588235294, 0.9176470588235294, 0.9490196078431372}{0.9176470588235294, 0.9176470588235294, 0.9490196078431372}{
    \includegraphics[width=.95\textwidth]{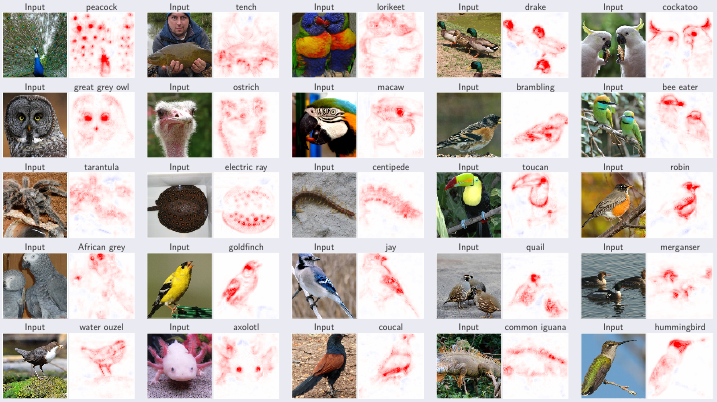}
    }
    \caption{The first 25 most confident classifications decomposed into the contributions from each spatial location, filtered to 1 image per class. Positive (negative) contributions for the ground truth class are shown in red (blue).}
    \label{fig:first_25}
    \vspace{-1em}
\end{figure*}

\begin{figure*}[t]
    \centering
    \fcolorbox[rgb]{0.9176470588235294, 0.9176470588235294, 0.9490196078431372}{0.9176470588235294, 0.9176470588235294, 0.9490196078431372}{
    \includegraphics[width=.95\textwidth]{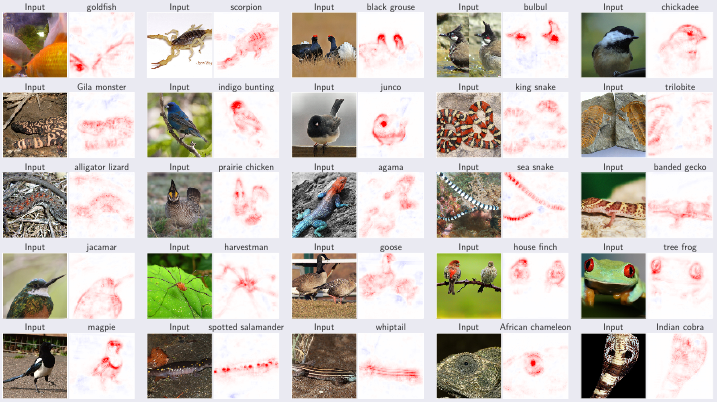}
    }
    \caption{The 26th to the 50th most confident classifications decomposed into the contributions from each spatial location, filtered to 1 image per class. Positive (negative) contributions for the ground truth class are shown in red (blue).}
    \label{fig:second_25}
\end{figure*}

\begin{figure*}[t]
    \centering
    \fcolorbox[rgb]{0.9176470588235294, 0.9176470588235294, 0.9490196078431372}{0.9176470588235294, 0.9176470588235294, 0.9490196078431372}{
    \includegraphics[width=.95\textwidth]{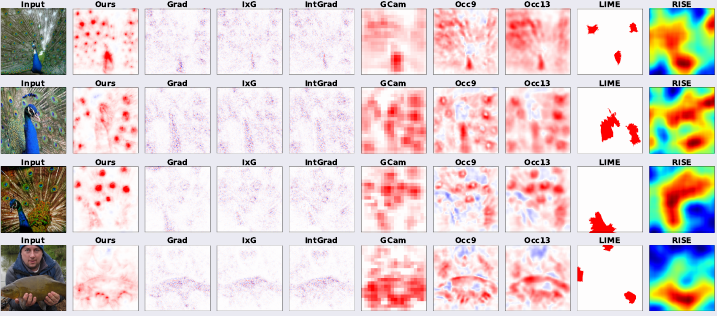}}
    \caption{Comparison between attribution methods for the 4 most confident classifications. We show positive importance attributions in red, negative attributions in blue; for RISE we use its default visualisation. Note that the model seems to align the weights well with the ornamental eyespots of the peacocks (also see the peacock in Fig.~\ref{fig:first_10_comp_2}) or the heads of the ducks in Fig.~\ref{fig:first_10_comp_2}. While the latter are also highlighted by GCam, the former constitute a structure that is too fine-grained for GCam to resolve properly.}
    \label{fig:first_10_comp_1}
\end{figure*}

\begin{figure*}[t]
    \centering
    \vspace{-1em}
    \fcolorbox[rgb]{0.9176470588235294, 0.9176470588235294, 0.9490196078431372}{0.9176470588235294, 0.9176470588235294, 0.9490196078431372}{
    \includegraphics[width=.95\textwidth]{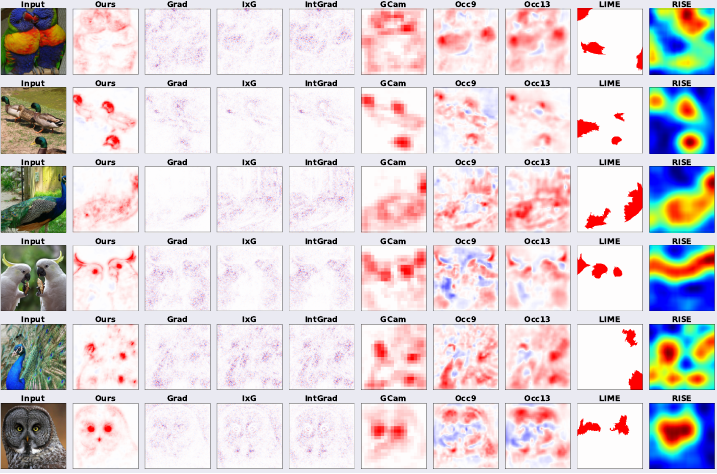}}
    \caption{Comparison between attribution methods for the 5th to the 10th most confident classifications. We show positive importance attributions in red, negative attributions in blue; for RISE we use its default visualisation. }
    \label{fig:first_10_comp_2}
    \vspace{-2em}
\end{figure*}

\myparagraph{Additional Imagenet examples.} In Figs.~\ref{fig:first_25} and~\ref{fig:second_25} we present additional qualitative examples of the model-inherent contribution maps. In particular, we show the decomposition of the model predictions into input contributions for 50 out of 100 classes; for each class, we show the most confidently classified image and show the classes in sorted order (by confidence). In Figs.~\ref{fig:first_10_comp_1} and~\ref{fig:first_10_comp_2}, we moreover show the contributions maps for the first 10 of the overall most confidently classified images next to the attribution maps from the post-hoc importance attribution methods for qualitative comparison. We note that GradCam consistently highlights very similar regions to the CoDA-Net contribution maps, but does so at a lower resolution. All contribution maps based on the CoDA-Net use the same linear color scale, which has been set to ($-v$, $v$) with $v$ the 99.75th percentile over all absolute values in the contributions maps shown in Figs.~\ref{fig:first_25} and~\ref{fig:second_25}. For reproducing the presented contribution maps and more, please visit \url{github.com/moboehle/CoDA-Nets}.

\begin{figure*}
    \centering
    \includegraphics[width=.95\textwidth]{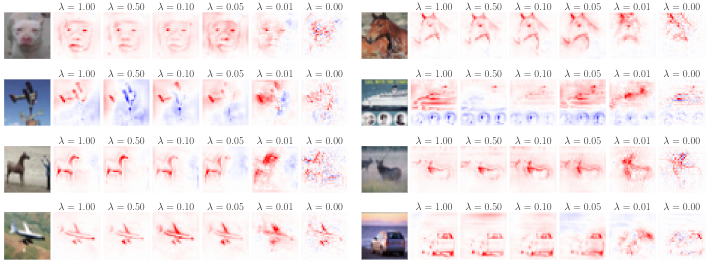}
    \caption{Visualising the qualitative effect of explicitly regularising the linear mapping $\mat w_{0\rightarrow L}$ on CIFAR-10. While the contribution maps without regularisation are noisy, they become sharper as soon as a regularisation is applied. }
    \label{fig:cifar10_regul}
    \vspace{-1em}
\end{figure*}

\myparagraph[0]{Regularising the linear mapping on CIFAR-10.}
As described in the main paper, the explicit representation of the model computations as a linear mapping, i.e., $$\hat{\vec y}(\vec x) = \mat w_{0\rightarrow L}(\vec x) \vec x\quad,$$
allows to explicitly regularise the linear mappings and thereby the model-inherent contribution maps. In Fig.~\ref{fig:cifar10_regul} we qualitatively show how the regularisation impacts the contribution maps. In particular, we see that without any regularisation the contribution maps are difficult to interpret. As soon as even a small regularisation is applied, however, the maps align well with the discriminative parts of the input image. 

\begin{figure*}[t]
    \centering
    \centering
    \includegraphics[width=.95\textwidth, trim=1em 1em 1em 1em, clip]{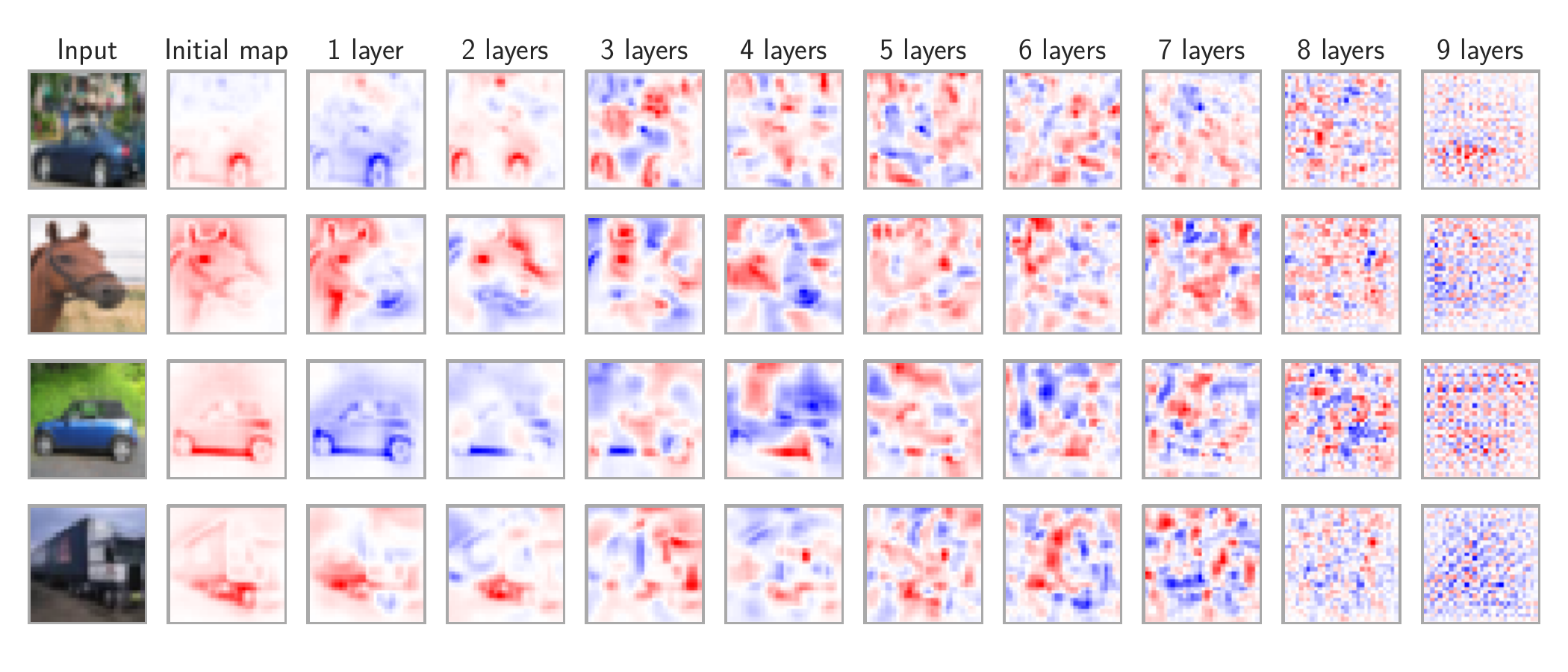}
    \caption{Sanity check experiment as in~\protect\citesupp{adebayo2018sanity}. If the importance attributions remain stable under parameter randomisation, they cannot be assumed to faithfully reflect the learnt parameters of the model. Since  the contribution maps get significantly perturbed when re-initialising layers
            from network output (left) to network input (right), the model-inherent contribution maps thus pass this sanity check.
            Positive contributions are shown in red, negative contributions in blue.
    }
    \label{fig:sanity_check}
\end{figure*}

\myparagraph{Sanity check.}
In~\citesupp{adebayo2018sanity}, the authors found that many commonly used methods for importance attribution are not \emph{model-faithful}, i.e., they do not reflect the learnt parameters of the model. In order to test this, they proposed to examine how the attributions change if the model parameters are randomised step by step. If the attributions remain stable under model randomisation, they cannot be assumed to explain a specific model, but rather reflect properties of the general architecture and the input data. In Fig.~\ref{fig:sanity_check}, we show how the model-inherent contribution maps behave when re-initialising the CoDA-Net layers one at a time to a random parameter setting, starting from the deepest layer. As can be seen, the contribution maps get significantly perturbed with every layer that is reset to random parameters; thus, the contribution maps pass this `sanity check' for attribution methods.
\clearpage

\myparagraph{Contribution maps of a piece-wise linear model.}
In Fig.~\ref{fig:resnet_maps} we show contribution maps obtained from different pre-trained ResNet architectures obtained from
    \url{https://github.com/akamaster/pytorch_resnet_cifar10}.
In particular, we visualise the `Input$\times$Gradient' method. Note that this yields contribution maps, since piece-wise linear models, such as the ResNets, produce input-dependent linear mappings, similar to the CoDA-Nets. These contribution maps, however, are rather noisy and do not reveal particularly relevant features.

\renewcommand{\putfig}[1]{%
\begin{subfigure}[b]{.45\textwidth}
                \centering
                \includegraphics[width=\textwidth, trim=1em 6em 1em 5em, clip]{supplement/resources/resnet_maps_#1.pdf}
            \end{subfigure}
}
\begin{figure*}
\centering
\begin{subfigure}[c]{\textwidth}
    \centering
    \putfig{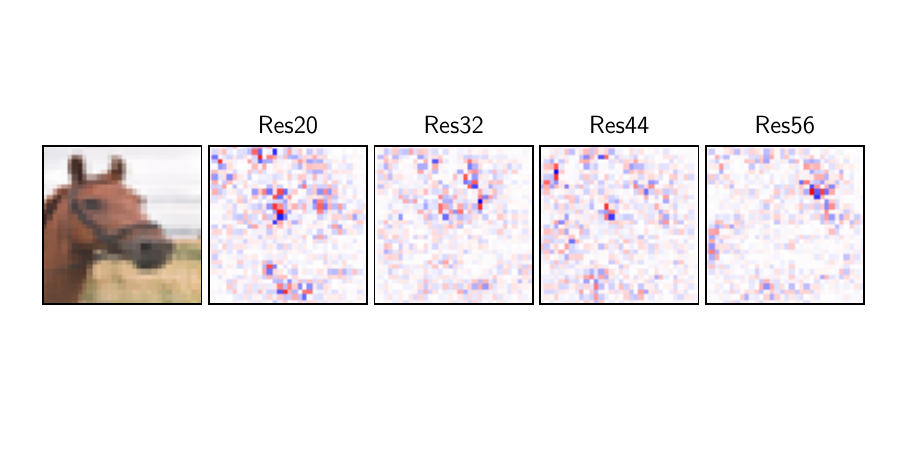}
    \putfig{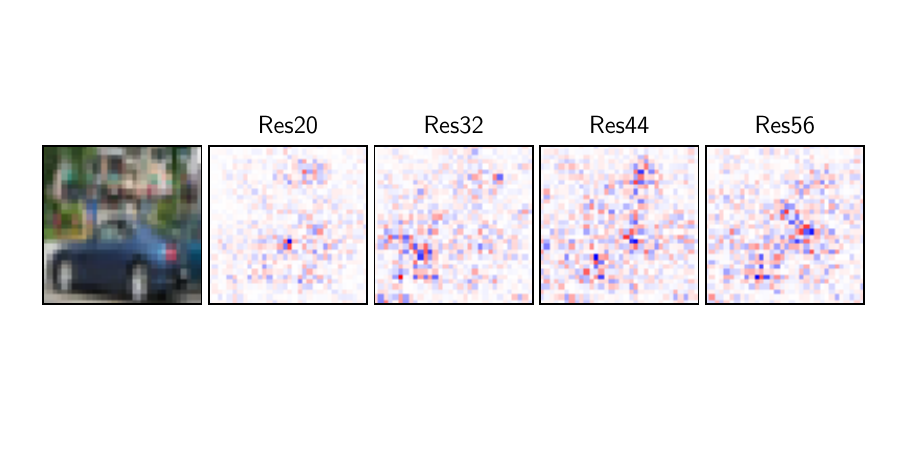}
    \putfig{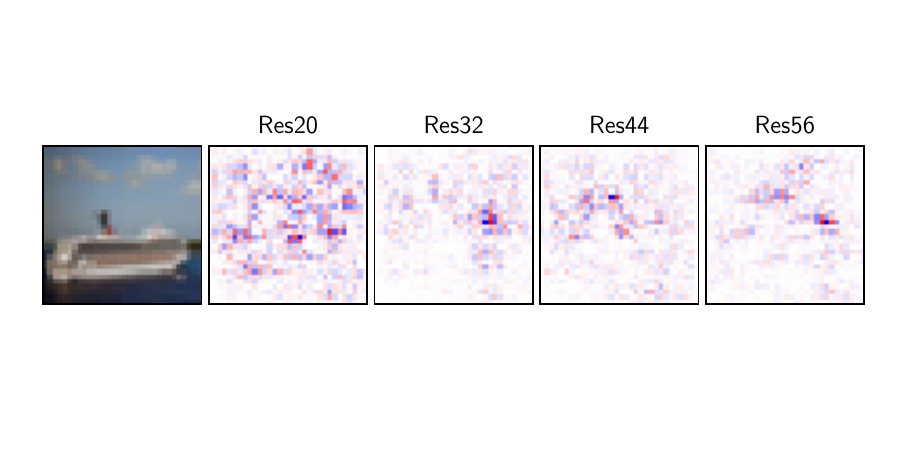}
    \putfig{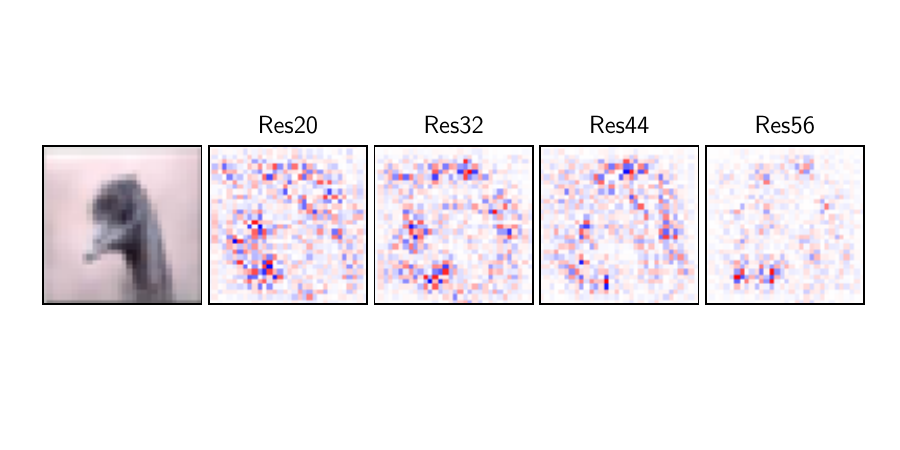}
    \putfig{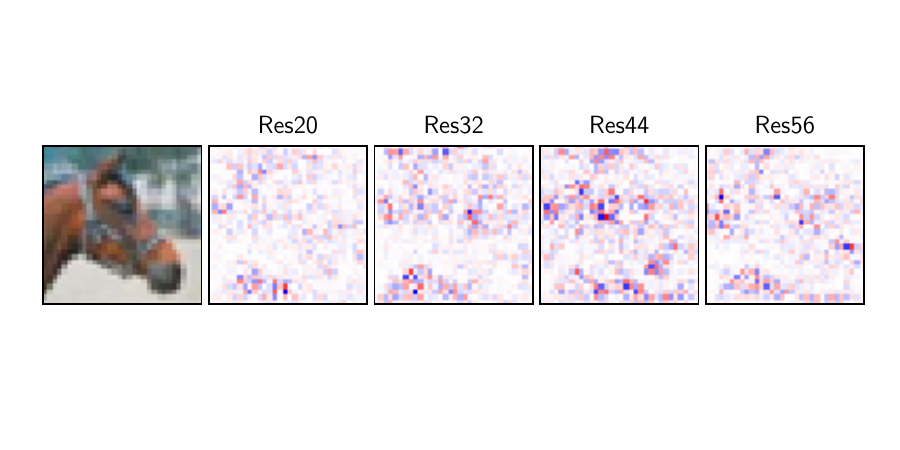}
    \putfig{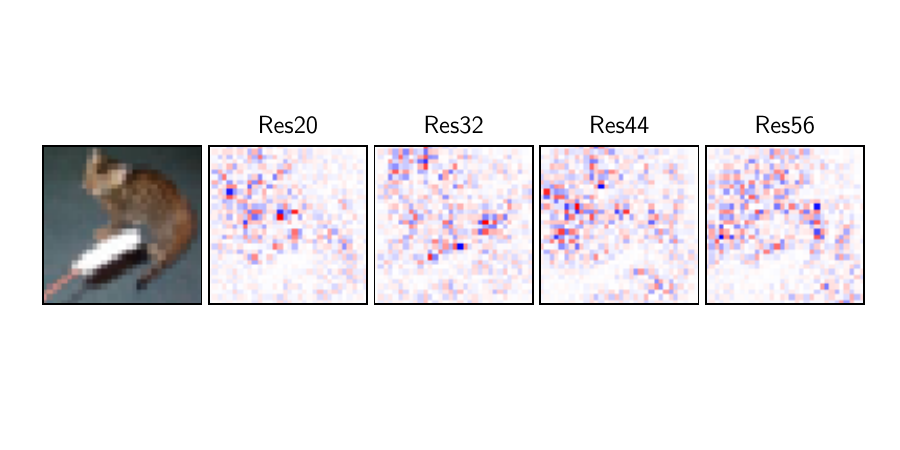}
    \putfig{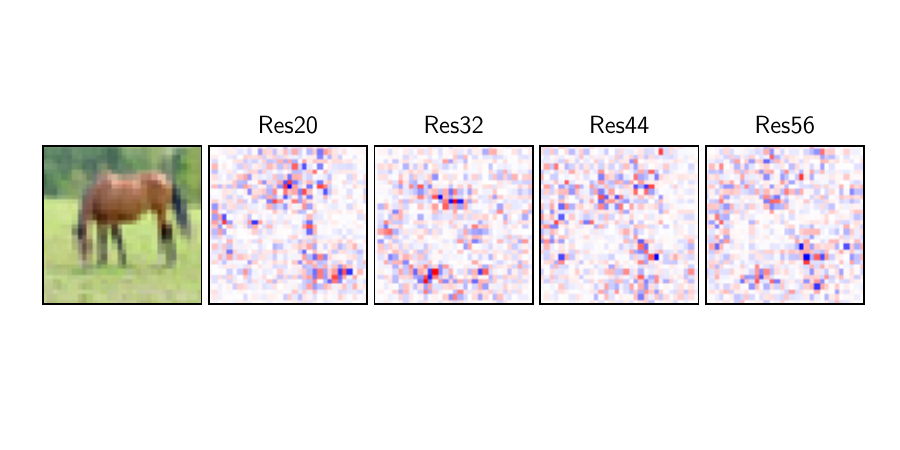}
    \putfig{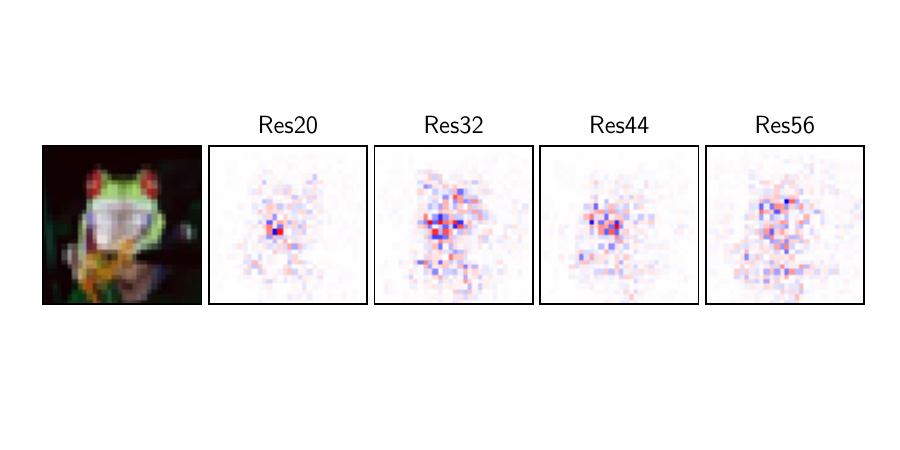}
    \vspace{-.5em}
\end{subfigure}
\caption{`Input$\times$Gradient' evaluated on different ResNets.
            Since ResNets are piece-wise linear functions, s.t. \mbox{$\vec y(\vec x) = \mat M(\vec x) \vec  x + \vec{b}(\vec x)$},
            this is the ResNet-based equivalent to the CoDA-Net contribution maps.
            }
\label{fig:resnet_maps}
\end{figure*}
\clearpage
\section{Additional quantitative results}
\label{sec:additional_quantitative}

\begin{figure}
\centering
    \begin{subfigure}[c]{\textwidth}
    \centering
    \begin{subfigure}[c]{0.45\textwidth}
    \centering
    \includegraphics[width=\textwidth, trim=0 1em 0 1em, clip]{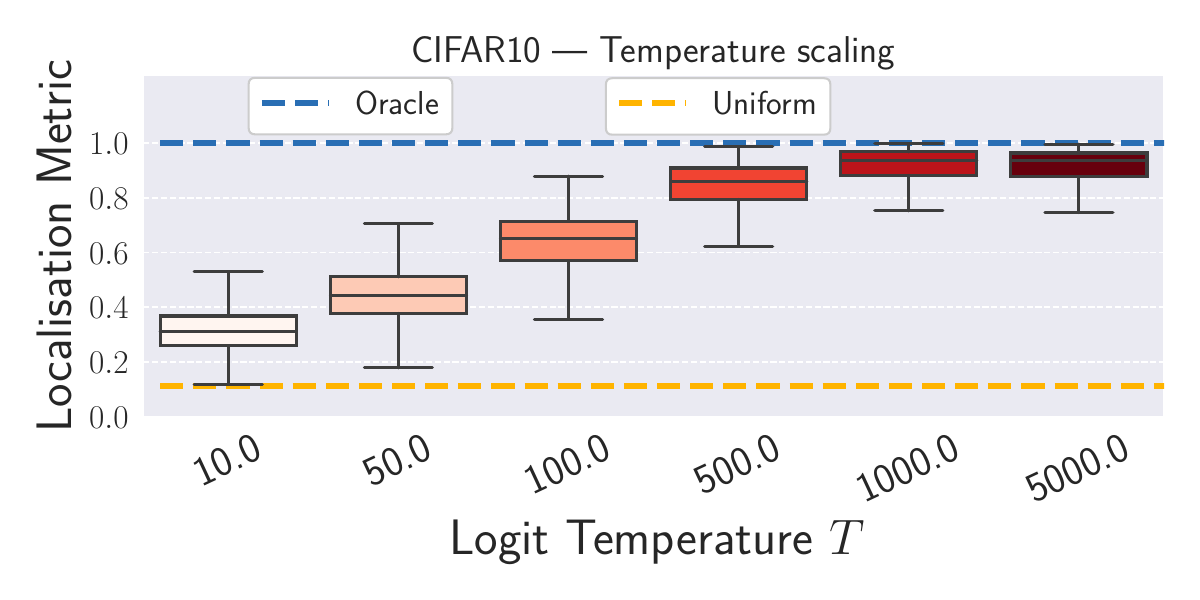}
    \end{subfigure}
    \begin{subfigure}[c]{0.45\textwidth}
    \centering
\includegraphics[width=\textwidth]{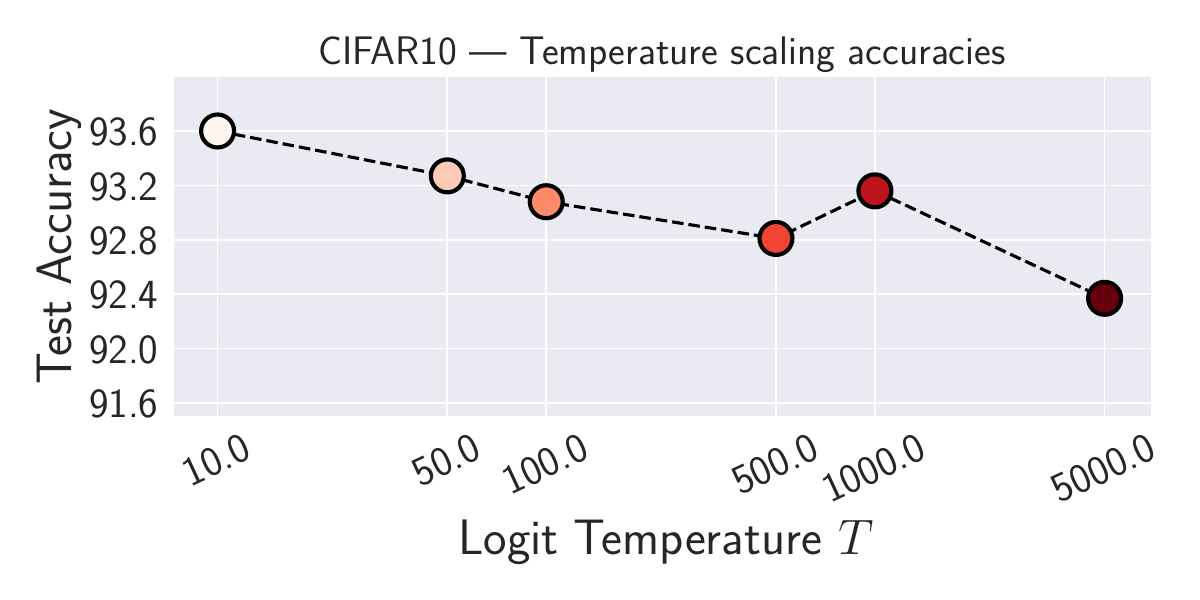}
    \end{subfigure}
    \vspace{.75em}
    \end{subfigure}
    \caption{\textbf{Left:} Localisation metric results for models trained with different temperatures $T$, same as in main paper (see Fig.~6, top right). 
    \textbf{Right:} Corresponding accuracies of the models on the CIFAR10 test set. There seems to be a trade-off between the interpretability and the accuracy of the models due to the regularising effect of $T$.}
    \label{fig:T_ablation}
\end{figure}

\myparagraph{Performance / interpretability trade-off.}
While the CoDA-Nets were observed to train and perform well over a wide range of choices for the logit temperature $T$, there seems to be a trade-off between the accuracies of the network and their interpretability---the implicit alignment regularisation comes at a cost. For example, in Fig.~\ref{fig:T_ablation}, we contrast the gain in interpretability (left, same figure as in main paper) with the corresponding accuracies (right). 

\begin{figure}
\centering
    \begin{subfigure}[c]{\textwidth}
    \centering
    \begin{subfigure}[c]{0.45\textwidth}
    \centering
    \includegraphics[width=\textwidth, trim=0 1em 0 1em, clip]{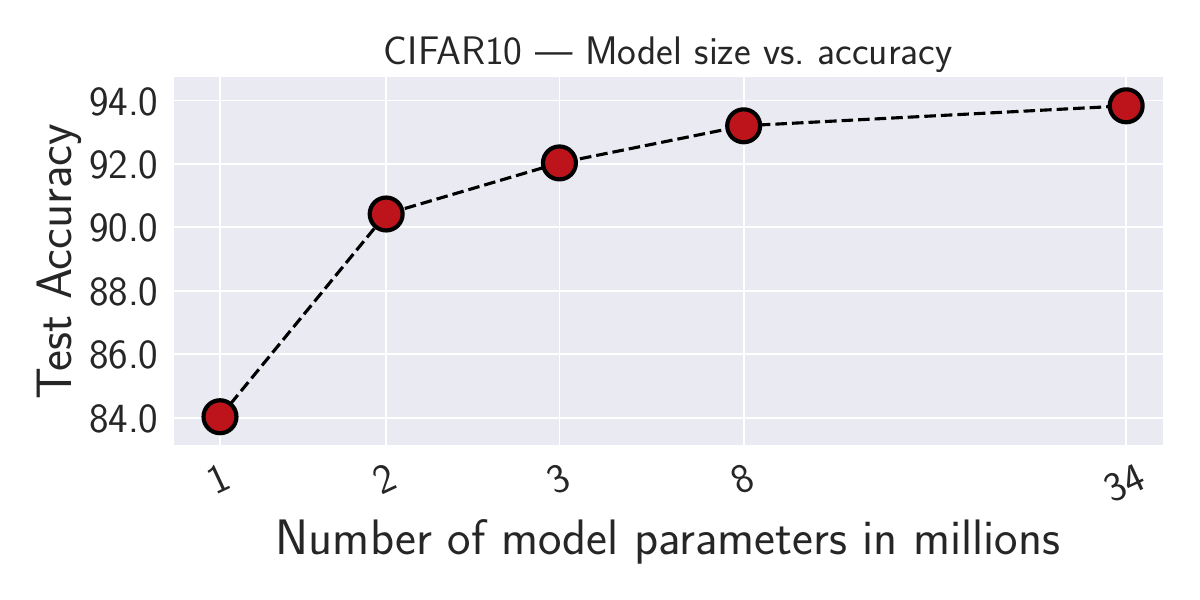}
    \end{subfigure}
    \vspace{.75em}
    \end{subfigure}
    \caption{Effect of scaling the rank of the DAUs in the CoDA-Nets on accuracy. Specifically, all ranks in the S-CoDA-SQ model presented in the main paper were scaled with the same factor, thereby changing the model size. The original model is shown with 8M parameters for comparison.}
    \label{fig:R_ablation}
\end{figure}

\myparagraph{Model size vs.~accuracy.}
Given the quadratic form in the DAUs (cf.~eq.~(1), $\approx \vec x^T\mat M \vec x$), the number of parameters per DAU scales quadratically with the input dimensions. In order to limit the model size, we decided to explicitly limit the rank of the DAUs by factorising the matrix $\mat M$ into $\mat a\mat b$ with $\mat a\in\mathbb{R}^{d\times r}$ and $\mat b\in\mathbb{R}^{r\times d}$.
While this allows to be more parameter efficient, it, of course, affects the modelling capacity of the DAUs. In Fig.~\ref{fig:R_ablation}, we present how the accuracy changes with the model size; for this, we scaled the ranks of all DAUs per layer with factors of 1/8, 1/4, 1/2, and 4.5 compared to the S-CoDA-SQ model presented in the main paper. This results in models with 1.1M, 2.0M, 4.0M, and 34.6M parameters respectively. For comparison, the original model is also included in Fig.~\ref{fig:R_ablation} (8M parameters).

\begin{figure}
\centering
    \begin{subfigure}[c]{\textwidth}
    \centering
    \begin{subfigure}[c]{0.45\textwidth}
    \centering
    \includegraphics[width=\textwidth, trim=0 1em 0 1em, clip]{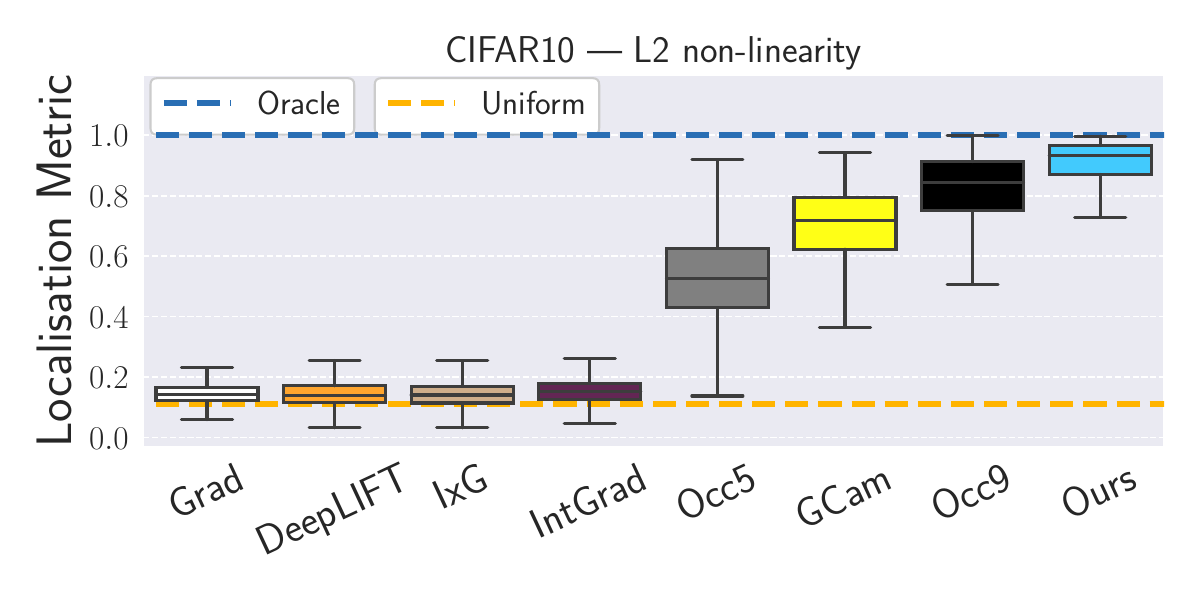}
    \end{subfigure}
    \begin{subfigure}[c]{0.45\textwidth}
    \centering
    \includegraphics[width=\textwidth, trim=0 1em 0 1em, clip]{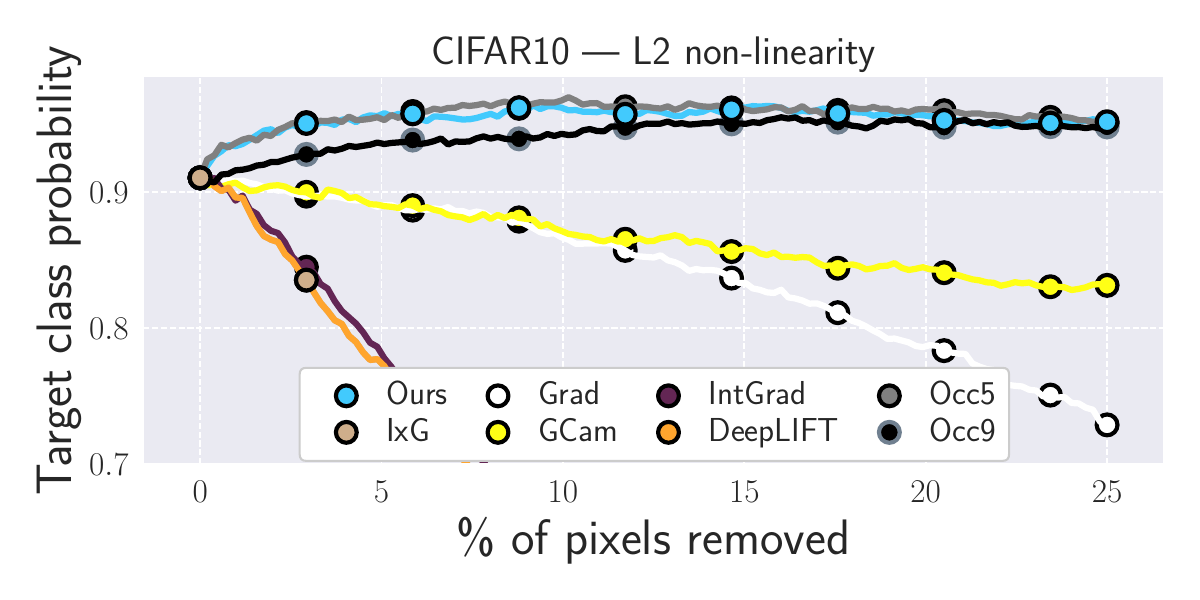}
    \end{subfigure}
    \caption{The quantitative results for a CoDA-Net trained with the L2 non-linearity are very similar to those of a model trained with the squashing non-linearity, see eq.~(2) in the main paper and compare with Fig.~6 (center column) in the main paper. In particular, we observe that the CoDA-Net outperforms the other methods under the localisation metric and achieves similar performance to the occlusion attribution method, which directly estimates the change in model prediction when removing a pixel.}
    \label{fig:l2_ablation}
    \vspace{.75em}
    \end{subfigure}
    \begin{subfigure}[c]{\textwidth}
    \centering
    \begin{subfigure}[c]{0.45\textwidth}
    \centering
    \includegraphics[width=\textwidth, trim=0 1em 0 1em, clip]{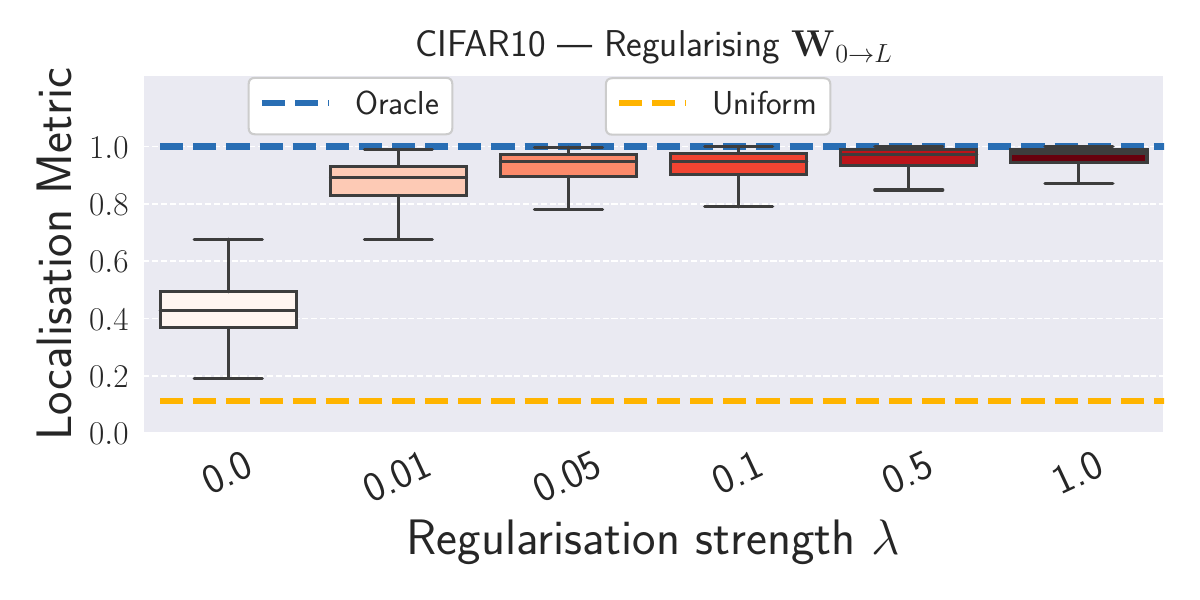}
    \end{subfigure}
    \begin{subfigure}[c]{0.45\textwidth}
    \centering
    \includegraphics[width=\textwidth, trim=0 1em 0 1em, clip]{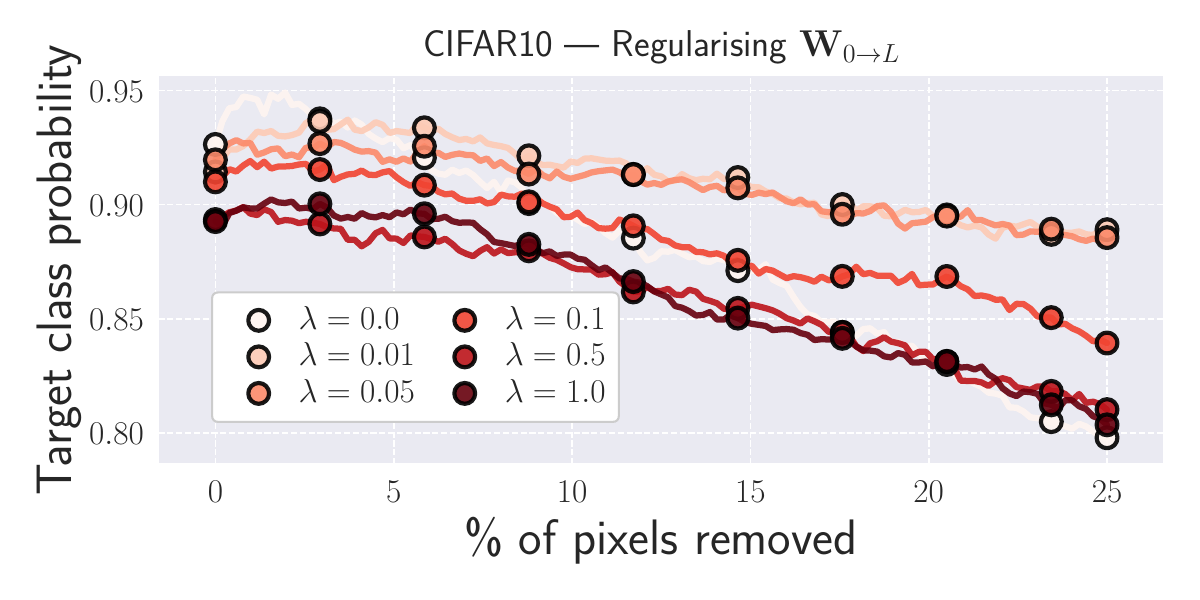}
    \end{subfigure}
    \caption{Similar to the effect of temperature scaling, cf.~Fig.~6 (right column) in the main paper, here we show quantitative results for different regularisation strengths $\lambda$. Similar to increasing the temperature, stronger regularisations also improve the localisation metric (left). While the pixel perturbation metric (right) also improves at first ($\lambda = 0.01$ and $\lambda = 0.05$), the models become less stable for stronger regularisation strengths. In comparison, we found that increasing the temperature also improves the performance of the models under this metric, see Fig.~6 (right column) in the main paper.}
    \label{fig:regul_ablation}
    \vspace{.75em}
    \end{subfigure}
    \caption{Quantitative results for two ablations: (a) using the L2 non-linearity and (b) increasing the regularisation of the linear mapping $\mat W_{0\rightarrow L}$, see~eqs.~(2) and (10) in the main paper respectively. Compare with Fig.~6 in the main paper.}
\end{figure}

\myparagraph{Results for L2 non-linearity.}
In Fig.~\ref{fig:l2_ablation} we show the results of evaluating the different methods for importance attribution on a model with the L2 non-linearity, see~eq.~(2) in the main paper. As can be seen, the results are very similar to those presented in the main paper in Fig.~6 (center column); in particular, the model-inherent contribution maps outperform the other methods under the localisation metric and are on par with the occlusion methods under the pixel removal metric; note, however, that the occlusion methods are a direct estimate of the behaviour under pixel removal and therefore expected to perform well under this metric.
\begin{figure}
\centering
    \begin{subfigure}[c]{\textwidth}
    \centering
    \begin{subfigure}[c]{0.45\textwidth}
    \centering
    \includegraphics[width=\textwidth]{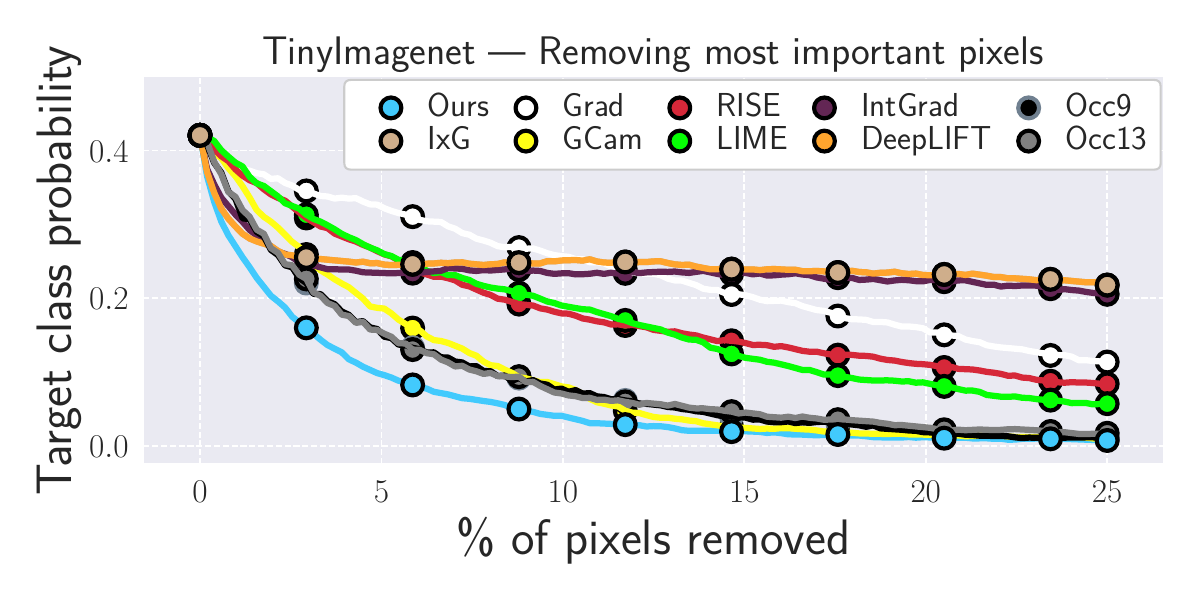}
    \end{subfigure}
    \caption{Results for the pixel removal metric when first removing the most important pixels. As can be seen, the ranking given by the model-inherent contribution maps seems to best reflect the pixel importance, since the confidence most rapidly drops when removing pixels according to this ranking.}
    \label{fig:most_important}
    \vspace{.75em}
    \end{subfigure}
    \begin{subfigure}[c]{\textwidth}
    \centering
    \begin{subfigure}[c]{0.45\textwidth}
    \centering
    \includegraphics[width=\textwidth, trim=0 1em 0 1em, clip]{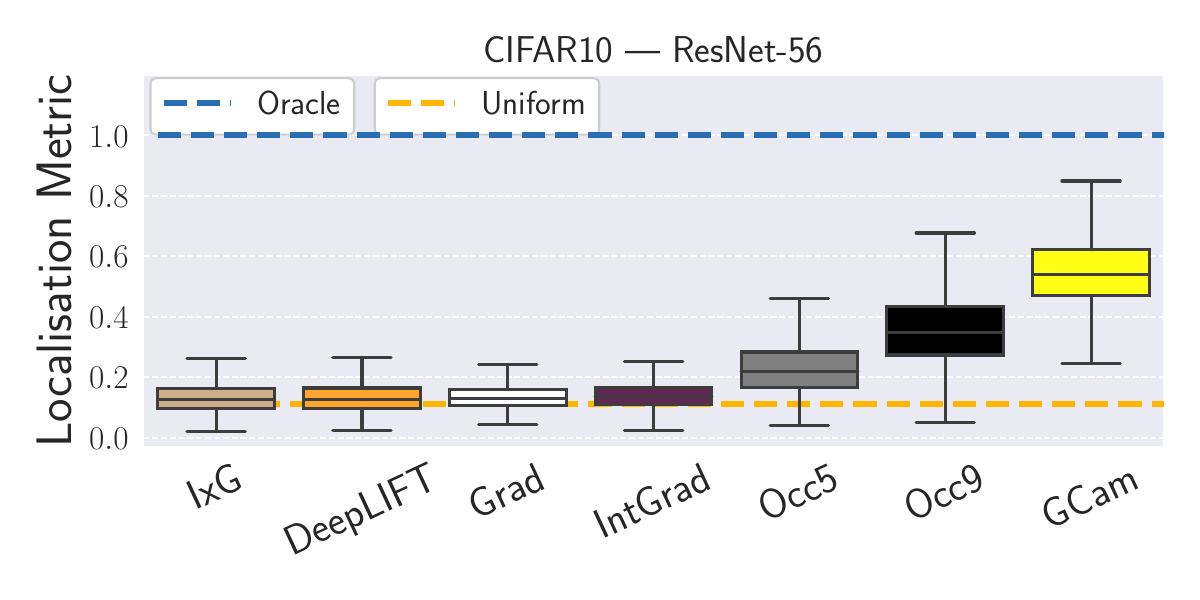}
    \end{subfigure}
    \begin{subfigure}[c]{0.45\textwidth}
    \centering
    \includegraphics[width=\textwidth, trim=0 1em 0 1em, clip]{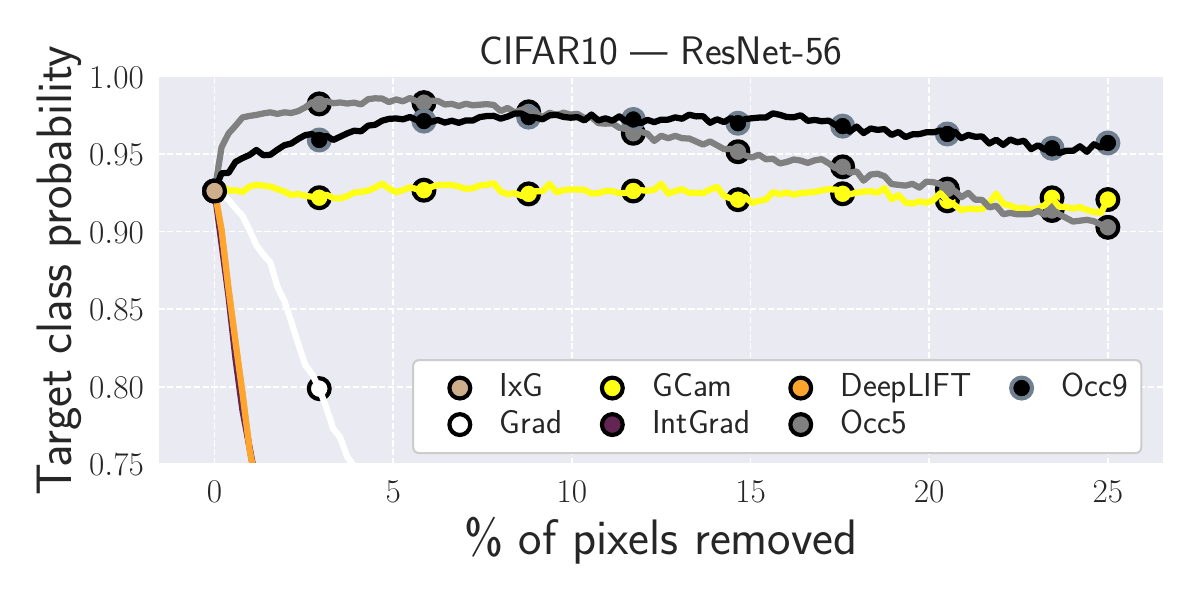}
    \end{subfigure}
    \caption{Quantitative results for methods for importance attribtution evaluated on a pre-trained ResNet-56 on CIFAR-10. Comparing these results to Fig.~6 (center column) in the main paper or Fig.~\ref{fig:l2_ablation}, it can be seen that the model-inherent contribution maps of the CoDA-Net are also strong when compared to importance attributions evaluated on a different model.}
    \label{fig:resnet_results}
    \vspace{.75em}
    \end{subfigure}
    \caption{In (a) we show the results of the pixel-removal metric when removing those pixels first that are considered the most important ones according to the importance attribution method. Moreover, in (b) we plot the quantitative results for the evaluation metrics of the importance attributions for a pre-trained ResNet-56.}
    \vspace{-.25em}
\end{figure}

\myparagraph{Regularisation of $\mat w_{0\rightarrow L}$.} In Fig.~\ref{fig:regul_ablation}, we show the results of the localisation metric and the pixel removal metric for models trained on CIFAR-10 with different regularisation strengths $\lambda$, see~eq.~(10) in the main paper. We note that the localisation metric benefits from an increase in the regularisation strength $\lambda$. For the pixel removal metric, we observe that while the models improve at first under this metric, a strong regularisation makes the predictions more brittle. 

\myparagraph{Removing the \emph{most important} pixels first.} In Fig.~\ref{fig:most_important}, we show the results of the pixel removal metric when removing the most important pixels first. As can be seen, the model-inherent contributions better predict the importance of pixels and outperform the other attribution methods under this metric; in particular, the confidence drops most rapidly when removing pixels according to the ranking given by the model-inherent contribution maps.

\myparagraph{Evaluating a pre-trained ResNet.} 
In order to establish a baseline for the performance of the different attribution methods on a classical neural network architecture, we additionally evaluated the attribution methods on a pre-trained ResNet and show the results in Fig.~\ref{fig:resnet_results}. Specifically, we rely on a publicly available 
    pre-trained ResNet-56 obtained from
    \mbox{\url{https://github.com/akamaster/pytorch_resnet_cifar10}}, which achieves a classification accuracy of $93.39\%$. 
The results show that the performance of the CoDA-Net-derived contribution maps is not only strong when comparing them to attribution methods evaluated \emph{on the same model}. Instead, they also perform well in comparison to those methods evaluated \emph{on a different model}. In particular, the model-inherent contribution maps of the CoDA-Net outperform attribution methods evaluated on the pre-trained ResNet-56 under the localisation metric. Further, only the occlusion attributions produce similarly strong pixel importance rankings for the ResNet; note, however, that the occlusion methods are a direct estimate of the behaviour under pixel removal and therefore expected to perform well under this metric.

\section{Implementation details}
\label{sec:training}
\subsection{Training and architecture details}
\myparagraph{Architectures.} The architectures used for the experiments in section 4 are given in Table~\ref{tbl:archs}.
All activation maps are padded with $(k - 1)/2$ zeros on each side, such that the spatial dimensions are only reduced by 
    the strides; here, $k$ refers to the kernel size.
As can be seen, the activation maps thus still have a spatial resolution after the last layer, which we further reduce with a global sum-pooling layer.
Note that global sum-pooling is just a linear layer with no trainable parameters and therefore still allows for linear decomposition.
The input itself consists of 6 channels, the image and its negative, as explained in section~3.4;
    hence, the first layer takes an input with 6 channels per pixel.

\myparagraph{Training details.} We use the pytorch library~\citesupp{pytorch} and optimise all networks with the Adam optimiser~\citesupp{kingma2014adam} with default values.
As for the loss function, we use the binary cross entropy loss to optimise class probabilities individually (as `one-vs-all' classifiers).
For all networks, we used a base learning rate of $2.5 \times10^{-4}$; for the Imagenet experiment, we employed learning rate warm-up and linearly increased the learning rate from $2.5 \times10^{-4}$ to $1 \times10^{-3}$ over the first 15 epochs.
Further, we trained for 200 epochs on CIFAR-10, for 100 epochs on TinyImagenet, and for 60 epochs on the Imagenet subset; we decreased the learning rate by a factor of 2 after every 60/30/20 epochs on CIFAR-10/TinyImagenet/Imagenet.
We used a batch size of 16, 128, and 64 for CIFAR-10, TinyImagenet, and Imagenet respectively. For the Imagenet subset, we additionally used RandAugment~\citesupp{cubuk2019randaugment} with parameters $n=2$ and $m=9$; for this, we relied on the publicly available implementation at \url{https://github.com/ildoonet/pytorch-randaugment} and followed their augmentation scheme. The qualitatively evaluated model (see Figs.~5,~8, and~A1-A4) for the Imagenet subset was trained with $T=1e5$ and achieved a top-1 accuracy of 76.5\%. For comparison, we trained several ResNet-50 models (taken from the pytorch library~\citesupp{pytorch}) with the exact same training procedure, i.e., batch size, learning rate, optimiser, augmentation, etc.). The best ResNet-50 out of 4 runs achieved 79.16\% top-1 accuracy\footnote{The best test accuracies per run are given by 79.16\%, 79.04\%, 78.86\%, and 78.7\% respectively.}, which outperforms the CoDA-Net but is nevertheless comparable. While it is surely possible to achieve better accuracies for both models, long training times for the CoDA-Nets have thus far prevented us from properly optimising the architectures both on the 100 classes subset, as well as on the full Imagenet dataset. In order to scale the CoDA-Net models to larger datasets, we believe it is important to first improve the model efficiency in future work.
Lastly, when regularising the matrix entries of $\mat W_{0\rightarrow L}$, see eq.~(11),
    we regularised the absolute values for the true class $c$, $\left[\mat M_{0\rightarrow L}\right]_c$,
    and a randomly sampled incorrect class per image.

\subsection{Convolutional Dynamic Alignment Units}
\label{subsec:CoDAUs}
In Algorithm \ref{alg:dlcconv2d}, we present the implementation of the convolutional DAUs (CoDAUs).
As can be seen, a Convolutional Dynamic Alignment Layer applies dynamic (input-dependent) filters to each of the patches extracted at different spatial locations. In detail, the Dynamic Alignment Units are implemented as two consecutive convolutions (lines 11 and 15), which are equivalent to first applying matrix $\mat B$ (line 24) and then $\mat A$ to each patch and adding a bias term $\vec b$ (line 29).
After applying the non-linearity (line 33), we obtain the dynamic weight vectors for CoDAUs as described in eq.~(1) in the main paper.
In particular, for every patch $\vec p_{hw}$ extracted at the spatial positions $hw$ in the input, we obtain the dynamic weight $\vec w_j(\vec p_{hw})$ for the $j$-th DAU as
\begin{align}
    \vec w_j(\vec p_{hw}) = g(\mat{a}_j\mat{b}\vec p_{hw} + \vec b_j)\; ;
\end{align}
note that the projection matrices $\mat b$ are thus shared between the DAUs.
These weights are then applied to the respective locations (line 41) to yield the outputs of the DAUs per spatial location.
As becomes apparent in line 41, the outputs are linear transformations (weighted sums) of the input and can be written as
\begin{align}
    \vec a_{l+1}(\vec a_l) = \mat W(\vec a_l) \vec a_l\quad ,
\end{align}
with $\vec a_l\in \mathbb R^{d}$ the vectorised input to layer $l$ and $\mat w\in\mathbb R^{f\times d}$ and $f$ the number of filters (DAUs).
The rows in matrix $\mat w$ correspond to exactly one filter (DAU) applied to exactly one patch $\vec p_{hw}$ and are non-zero only at those positions that correspond to this specific patch in the input.

\subsection{Attribution methods}
\label{subsec:att_details}
In section~4.2, we compare the model-inherent contribution maps of the CoDA-Nets to those of the following methods for importance attribution:
 the gradient of the class logits with respect to the input image~\citesupp{baehrens2010explain} (Grad), `Input$\times$Gradient' (IxG, cf.~\citesupp{adebayo2018sanity}), GradCam~\citesupp{selvaraju2017grad} (GCam), Integrated Gradients~\citesupp{sundararajan2017axiomatic} (IntG), DeepLIFT~\citesupp{shrikumar2017deeplift},
    several occlusion sensitivities (Occ-K, with K the size of the occlusion patch)~\citesupp{zeiler2014visualizing},
    RISE~\citesupp{petsiuk2018rise}, and LIME~\citesupp{lime}.

For RISE and LIME, we relied on the official implementations available at \url{https://github.com/eclique/RISE} and 
\url{https://github.com/marcotcr/lime} respectively. For RISE, we generated 6000 masks with parameters $s=6$ and $p_1=0.1$.
For LIME, we evaluated on 256 samples per image and used the top 3 features for the attribution;
    for the segmentation, we also used the default parameters, namely `quickshift' with $max\_dist=200$, $ratio=0.2$, and a kernel size of 4.

For Grad, GCam, IxG, IntG, DeepLIFT, and the occlusion sensitivities,
    we relied on the publicly available pytorch library `captum' (\url{https://github.com/pytorch/captum}).
GCam was used on the last activation map before global sum-pooling.
The occlusion sensitivities were used with 
    strides of 2 on CIFAR-10 and strides of 4 for TinyImagenet.
Finally, for IntG we used 20 steps for the integral approximation.

\subsection{Evaluation metrics}
\label{subsec:metric_details}
In section~4.2, we evaluated the attribution methods against 2 quantitative 
metrics: 
\colornum{(1)} the adapted \emph{pointing game}~\citesupp{zhang2018top} and
\colornum{(2)} the prediction stability under removing the \emph{least important pixels} as in~\citesupp{srinivas2019full}. In section~\ref{sec:additional_quantitative}, we further show results for removing the \emph{most important pixels} first.

For \colornum{(1)}, we constructed 500 (250) $3\times3$ multi-images for CIFAR-10 (TinyImagenet); for an example with $2\times2$, see Fig.~7 in the main paper.
In each of these multi-images, every class occurred at most once. As stated in section~4.2,
    we measured the fraction of positive contributions falling inside the correct mini-image. Further, the images were sorted according to their confidence for each of the classes. For every multi-image, a random set of classes was sampled. For each of the classes, we included the most confidently classified class image in the multi-image that had not been used yet in previous multi-images.

For \colornum{(2)}, we followed~\citesupp{srinivas2019full} and successively replaced one pixel at a time by [0, 0, 0, 0, 0, 0],
    until up to 25\% of the image were removed. The pixels were removed in order, sorted by their assigned importance.
\begin{table*}
    \newcommand{\rot}[1]{%
\rotatebox[origin=c]{90}{#1}}
\footnotesize
\centering
    \begin{tabular}{c|c|c|c|c|c|c}
         \textbf{Network} &
         \textbf{Input dimensions} &
         \textbf{Layer} &
         \textbf{Number of DAUs} &
         \textbf{Rank of $\mat{AB}$} &
         \textbf{Kernel size} &
         \textbf{Stride} 
         \\[.5em]
         \hline
         & & & & & &   \\[-.5em]
         \multirow{9}{*}{\rot{S-CoDA}}&
         \multirow{9}{*}{{$6\times 32\times 32$}}
         & 1 & 16 &  32 & 3 & 1 \\
         && 2 & 16 &  32 & 3 & 1 \\
         && 3 & 32 &  64 & 3 & 2 \\
         && 4 & 32 &  64 & 3 & 1 \\
         && 5 & 32 &  64 & 3 & 1 \\
         && 6 & 64 &  {64} & 3 & 2 \\
         && 7 & 64 &  {64} & 3 & 1 \\
         && 8 & 64 &  {64} & 3 & 1 \\
         && 9 & 10 &  {64} & 1 & 1 \\[.5em]
         \hline
         & & & & & &  \\[-.5em]
         \multirow{9}{*}{\rot{M-CoDA}}&
         \multirow{9}{*}{{$6\times 32\times 32$}}
         & 1 & 16 &  64 & 3 & 1 \\
         && 2 & 16 &  64 & 3 & 1 \\
         && 3 & 32 &  128 & 3 & 2 \\
         && 4 & 32 &  128 & 3 & 1 \\
         && 5 & 32 &  128 & 3 & 1 \\
         && 6 & 64 & {256} & 3 & 2 \\
         && 7 & 64 & {256} & 3 & 1 \\
         && 8 & 64 & {256} & 3 & 1 \\
         && 9 & 10 & {256} & 1 & 1 \\[.5em]
         \hline
         & & & & & &  \\[-.5em]
         \multirow{9}{*}{\rot{L-CoDA}}&
         \multirow{9}{*}{$6\times 240\times 240$}
         & 1 & 16 &   {64} & 7 & 3 \\
         && 2 & 32 &   {64} & 3 & 1 \\
         && 3 & 32 &  {64} & 3 & 1 \\
         && 4 & 64 &  {128} & 3 & 2 \\
         && 5 & 64 &  {128} & 3 & 1 \\
         && 6 & 64 &  {128} & 3 & 1 \\
         && 7 & 64 &  {256} & 3 & 2 \\
         && 8 & 64 &  {256} & 3 & 1 \\
         && 9 & 100 &  {256} & 3 & 1 \\[.5em]
         \hline
         & & & & & &  \\[-.5em]
         \multirow{9}{*}{\rot{XL-CoDA}}&
         \multirow{9}{*}{$6\times 64\times 64$}
         & 1 & 16 &   {64} & 5 & 1 \\
         && 2 & 32 &   {64} & 3 & 1 \\
         && 3 & 32 &  {128} & 3 & 2 \\
         && 4 & 64 &  {128} & 3 & 1 \\
         && 5 & 64 &  {128} & 3 & 1 \\
         && 6 & 64 &  {256} & 3 & 2 \\
         && 7 & 64 &  {256} & 3 & 1 \\
         && 8 & 64 &  {256} & 3 & 1 \\
         && 9 & 200 &  {256} & 3 & 2 \\
    \end{tabular}
\normalsize
    \vspace{1em}
    \caption{Architecture details for the experiments described in section 4.}
    \label{tbl:archs}
\end{table*}
{\centering
\RestyleAlgo{ruled}\LinesNumbered\setlength{\algomargin}{1.5em}
\begin{algorithm*}[htpb]
\caption{Implementation of a Convolutional Dynamic Alignment Layer}
  \label{alg:dlcconv2d}
    \setstretch{1.25}
  \DontPrintSemicolon
  \newcommand{\mycomment}[1]{{\color[RGB]{112, 128, 144}\textit{\# #1}}\;}
  \newcommand{\self}{{\bf\color[RGB]{0,24,128}{self}}}
  \newcommand{\pykey}[1]{{\bf\color[RGB]{0,128,24}{#1}}}
  \newcommand{\pyword}[1]{{\bf\color[RGB]{197,117,50}{#1}}}
  \SetKwFunction{FMain}{DAUConv2d(nn.Module)}
  \SetKwFunction{Finit}{\_\_init\_\_}
  \SetKwFunction{Ffwd}{forward}
  \SetKwProg{Fn}{\pykey{class}}{:}{}
  \SetKwProg{Imp}{}{}{}
  \SetKwProg{Df}{\pykey{def}}{:}{}
  \Imp{\normalfont \pykey{from} torch \pykey{import} nn}{}\vspace{-.25em}
  \Imp{\normalfont \pykey{import} torch.nn.functional \pykey{as} F}{}\vspace{-.6em}\;\vspace{-.25em}
  \Fn{\FMain}{\;
    \Df{\Finit{\textit{\self, in\_channels, out\_channels, rank, kernel\_size, stride, padding, act\_func}}}{
        \mycomment{act\_func: non-linearity for scaling the weights. E.g., L2 or SQ.}
        \mycomment{out\_channels: Number of convolutional DAUs for this layer.}
        \mycomment{rank: Rank of the matrix $\mat{AB}$.}
        \mycomment{`dim\_reduction' applies matrix $\mat b$.}
        \self.dim\_reduction = nn.Conv2d(\textit{in\_channels, rank, kernel\_size, stride, padding}, bias=\pyword{False})\;
        \mycomment{Total dimensionality of a single patch}
        \self.patch\_dim = in\_channels $\ast$ kernel\_size $\ast$ kernel\_size\;
        \mycomment{`weightings' applies matrix $\mat a$ and adds bias $\vec b$.}
        \self.weightings = nn.Conv2d(\textit{rank, out\_channels $\ast$ \self.patch\_dim , kernel\_size=1, bias=\pyword{True}})\;
        \self.act\_func = act\_func\;
        \self.out\_channels = out\_channels\;
        \self.kernel\_size = kernel\_size\;
        \self.stride = stride\;
        \self.padding = padding\;
        }\;
        \Df{\Ffwd{\textit{\self, in\_tensor}}}{
        \mycomment{Project to lower dimensional representation, i.e., apply matrix $\mat B$. This yields $\mat B\vec p$ for every patch $\vec p$.}
        reduced = \self.dim\_reduction(in\_tensor)\;
        \mycomment{Get new spatial size height h and width w}
        h, w = reduced.shape[--2:]\;
        batch\_size = in\_tensor.shape[0]\;
        \mycomment{Apply matrix $\mat A$ and add bias $\vec b$,
        yielding $\mat a\mat b\vec p + \vec b$ for every patch $\vec p$.}
        weights = \self.weightings(reduced)\;
        \mycomment{Reshape for every location to size \textit{patch\_dim}$\times$out\_channels}
        weights = weights.view(batch\_size, \self.patch\_dim, out\_channels, h, w)\;
        \mycomment{Apply non-linearity to the weight vectors, yielding $\vec w(\vec p) = g(\mat{ab} \vec p + \vec b$) as in eq.~(1) for every patch $\vec p$.}
        weights = \self.act\_func(weights, dim=1)\;
        \mycomment{Extract patches from the input to apply dynamic weights to patches.}
        patches = F.unfold(in\_tensor, \self.kernel\_size, padding=\self.padding, stride=\self.stride)\;
        \mycomment{Reshape for applying weights.}
        patches = patches.view(batch\_size, \self.patch\_dim, 1, h, w)\;
        \mycomment{Apply the weights to the patches.}
        \mycomment{As can be seen, the output is just a weighted combination of the input, i.e., a linear transformation.}
        \mycomment{The output can thus be written as $\vec o = \mat W(\vec x)\vec x$.}
        \KwRet\ (patches $\ast$ weights).sum(1)}
        }\end{algorithm*}
}

\newpage
\section{Low-rank matrix approximations}
\label{sec:low_rank_matrices}
In the following, we first introduce the standard formulation of the low-rank matrix approximation problem and present the well-known solution via singular value decomposition.
We then introduce an additional constraint to this standard formulation and show that at their optimum (maximal average output), the DAUs solve this constrained approximation problem.

\myparagraph{Low-rank approximation problem.}
Given a data matrix\footnote{In the context of our work, we can think of  $\mat m$ as storing $n$ data samples of dimensionality $m$; e.g., when considering images, each column might correspond to one vectorised image from a dataset of $n$ images. The low-rank approximation problem aims to approximate this dataset with $r$ independent variables (dimensionality reduction).} $\mat M \in \mathbb R^{m\times n}$, the low-rank approximation problem is typically formulated as $\min_{\mat M'}(||\mat M-\mat M'||_F)$ with $\textit{rank}(\mat M') \leq  r$ and $||\cdot||_F$ the Frobenius norm.
The Frobenius norm can be written as 
a sum over the column differences $||\mat \Delta_i||_F^2 = ||\vec m_i||_2^2 + ||\vec m'_i||^2_2 - 2\vec m^T_i\vec m_i'$ with $\vec m_i, \vec m_i'\in\mathbb R^{m}$ the $i$-th columns of the matrices $\mat m$ and $\mat m'$ respectively.
Note that since the $\vec m_i$ are fixed (they just correspond to the fixed input matrix $\mat m$), the optimisation objective is equivalent to
\begin{align}
	\label{eq:low_r_recon}
	&\min \textstyle\sum_i ||||\vec m'_i||^2_2 - 2\vec m^T_i\vec m_i'=\\
	\label{eq:svd2}
	&\min\textstyle\sum_i ||||\vec m'_i||^2_2 - 2||\vec m_i||||\vec m'_i||\cos(\alpha_i)\quad,
\end{align}
 with $\alpha_i$ the angle between $\vec m_i$ and $\vec m_i'$. Hence, optimising the Frobenius norm of the difference matrix finds a trade-off between aligning the directions (angles) of the columns of $\mat m$ and $\mat m'$ and approximating the correct norms of the original columns in \tmat m.
The optimal solution for this problem is given by singular value decomposition (SVD)~\citesupp{eckart1936approximation} of \tmat m, and the optimal $\mat m'$ can be written as 
\begin{align}
\label{eq:optimal_svd}
\mat m' = \mat U_r \mat U^T_r\mat m =\mat U\mat \Sigma_r\mat v^T\quad,
\end{align}
where $\mat U_r$ is the matrix of left singular vectors of $\mat m$ up to the $r$-th vector\footnote{A short proof of the equality to the right can be found at the end of this section.}; the remaining vectors are replaced by zero-vectors. Note that the first formulation of the solution ($\mat u_r\mat u_r^T\mat m $) 
emphasises a property that we will encounter again for the DAUs. In particular, this formulation highlights the fact that the optimal matrix $\mat m'$ is given by reconstructing the individual data points from the $r$-dimensional eigenspace spanned by the eigenvectors of the data points; i.e., each column $\vec m_i'$ can be calculated as $\mat u_r\mat u_r^T\vec m_i$.
The right hand side of the equation shows the conventional SVD-notation of the low-rank approximation solution, with $\mat u\in \mathbb R^{m\times m}, \mat \Sigma_r \in \mathbb R^{m\times n}\, \mat V \in \mathbb R^{n\times n}$ and $\mat \Sigma_r$ being a rectangular diagonal matrix with the first $r$ singular values as non-zero entries.

\myparagraph{Constrained low-rank approximations}
Now we diverge from the standard low-rank approximation formulation and impose an additional constraint on $\mat m'$ (red box):
\newcommand{\mycbox}[1]{%
\colorlet{oldcolor}{.}
  {\color[RGB]{150, 0, 0}
  \boxed{\color{oldcolor}#1}}
}
\begin{align}
    &\textstyle\min_{\mat M'}(||\mat M-\mat M'||_F)\;\text{,}\\
    \text{s.t.}\quad \textit{rank}(&\mat m')  \leq r \;\mycbox{\land\; ||\vec m_i'||_2^2 = 1 \;\forall i}\;\textbf{.}
    \label{eq:constraint}
\end{align}

When combining eqs.~\eqref{eq:svd2} and \eqref{eq:constraint}, it becomes clear that the optimisation problem can now be formulated as 
\begin{align}
    \textstyle\max\sum_i{\vec m}_i^T\vec m_i' = \max\sum_i||\vec m_i||_2\cos \left(\angle(\vec m_i,\vec m_i')\right)\; ,
    \label{eq:new_goal2}
\end{align}
since the norm of $\vec m_i'$ is fixed and the addition of or multiplication with a fixed constant does not change the location of the optima.

As a corollary, we note that \emph{any} matrix $\mat m'$ with a maximum rank $r$ that maximises eq.~\eqref{eq:new_goal2} gives a solution to the constrained approximation problem, \emph{independent} of the norm of its columns: the columns can be normalised \emph{post-hoc} to fulfill the new constraint, since rescaling does not change neither the rank of the matrix, nor the score of the objective function in eq.~\eqref{eq:new_goal2}.

Further, we note that the unnormalised optimal matrix $\widetilde{\mat m}'$ can be factorised as 
\begin{align}
    \label{eq:factorise}
    &\widetilde{\mat m}' = \mat a\mat b\mat m \; ,\\
    \text{s.t.}\quad  &\mat m' = G(\widetilde{\mat m}') = G(\mat{ABM})
\end{align}
with $\mat A\in \mathbb R^{m\times r}$ and $\mat b \in \mathbb R^{r\times n}$, similar to the solution to the unconstrained matrix approximation problem in eq.~\eqref{eq:optimal_svd}.
Here, $G$ is a function that normalises the columns of a matrix to unit norm (as discussed in the previous paragraph, in order to fulfill the norm constraint, we can normalise the columns \emph{post-hoc}).

Finally, this allows us to rewrite the objective function in eq.~\eqref{eq:new_goal2} as
\begin{align}
    \max \sum_i g(\mat{AB}\vec m_i)\vec m_i =\max \sum_i \vec w(\vec m_i)\vec m_i
\end{align}
with $g(\vec v)$ normalising its input vector to unit norm. 
Comparing this with eq.~(1) in the main paper, we see that this is equivalent to maximising a DAU without bias term for maximal output over the $n$ columns in $\mat m$ and therefore, at their optimum, the DAUs solve the constrained low-rank matrix approximation problem in eq.~\eqref{eq:constraint}. In particular, the DAUs achieve this by encoding common inputs in the eigenvectors of $\mat{ab}$, which allows for an optimal angular reconstruction of the inputs, similar to reconstructing from the first PCA components in a PCA decomposition.
The main difference to PCA is that PCA yields optimal reconstructions under the L2-norm, whereas the DAUs yield optimal angular reconstructions.

\myparagraph{Proof of $\mat U_r\mat U^T_r \mat m = \mat u \mat\Sigma_r\mat v^T$.}\\
In order to see that this equality holds, we first write matrix $\mat m$ as
\begin{align}
 \mat m = \mat U\mat \Sigma\mat v^T \quad (\text{SVD-form})
\end{align}
Now, we multiply both sides with $\mat U_r\mat U_r^T$ from the left:
\begin{align}
  \mat u_r \mat U^T_r  \mat M  &= \mat u_r \mat u_r^T\mat U\mat \Sigma\mat v^T \\
 \xrightarrow{\mat U_r^T \mat U = \mat I_r}
 \mat u_r \mat U^T_r  \mat M  &= \mat u_r \mat I_r\mat \Sigma\mat v^T \\
 \Rightarrow \mat u_r \mat U^T_r  \mat M  &= \mat U\mat \Sigma_r\mat v^T\quad \qed.
\end{align}
Here, we made use of the fact that $\mat u$ is an orthogonal matrix and therefore $\mat U^T\mat u =\mat I$; since we only use the first $r$ eigenvectors, we obtain the truncated identity matrix $\mat i_r$ when multiplying $\mat u^T_r\mat u$.
Further, in the last line, we used that $\mat u\mat\Sigma_r = \mat U_r\mat \Sigma_r$.

\section{Comparison to capsule networks}
\label{sec:capsule_comparison}
In order to discuss the relationship to capsule networks, in section~\ref{subsec:capsule_diff_notation} we will first rewrite the classical capsule formulation in
    `Dynamic Routing Between Capsules' by Sabour et al.~\citesupp{sabour2017dynamic} 
    to mitigate the notational differences between their work and our work.
In section~\ref{subsec:capsule_diff_usage} we then show that while capsules and Dynamic Alignment Units share some computations, there are several important differences that we summarise in Table~\ref{tbl:caps}.
\begin{table}[th]
    \vspace{.5em}
    \centering
    \begin{tabular}{r | c c}
        &\small\textbf{Classical capsules$\,\,$} &
        \small\textbf{Dynamic Alignment Units} \\\cline{1-3}&&\\[-.1em]
         \small\textbf{Non-Lin. $g$} &
         \small
         SQ&
         \small
         SQ, L2, ...\\[.5em]
         \small\textbf{Activations} &
         \small
         $ g\left(\mat v \vec x\right)$&
         \small$
         g\left(\mat{AB}\vec x+ \vec b\right)$\\[.5em]
         \small\textbf{Routing} &
         \small\textbf{yes} &
         \small\textbf{no}\\[.5em]
         \small\textbf{Low-rank} &
         \small\textbf{no} &
         \small\textbf{yes}\\[.5em]
         \small\textbf{Output} &
         $\text{CAP}(\vec x)$ &
         $\text{CAP}(\vec x)^T\vec x$
    \end{tabular}
    \caption{Comparison between capsules and Dynamic Alignment Units (DAUs). Importantly, DAUs produce a linear transformation of the input by multiplying the `capsule activations' with the input (see `Output') and allow for constraining the rank of the transformation. The dynamic weights in the DAUs can be seen as the activations of a capsule, s.t.~$\vec w(\vec x)=\text{CAP}(\vec x)$, see `Output' in the table.}
    \label{tbl:caps}
    \vspace{-.5em}
\end{table}
\subsection{Reformulating capsules}
\label{subsec:capsule_diff_notation}
In this subsection, we will show that the classical capsule formulation (eq.~\eqref{eq:sabour_caps1}), in which input capsules `vote' for the activations of an output capsule, can be written as a simple linear transformation $\vec S = \mat v \vec x$ if just one iteration of the dynamic routing algorithm is applied; here, $\vec s$ is a vector containing the activations of the output capsule, $\mat v$ stores the `votes' of the input capsules to an output capsule, and $\vec x$ is a vector containing the activations of all input capsules. In the following, we will start from how capsules are formulated in~\citesupp{sabour2017dynamic} and rewrite this formulation step by step.

In~\citesupp{sabour2017dynamic} eq.~(2), the authors calculate the activations $\vec s$ of a capsule\footnote{
    As we only discuss a single output capsule, we omitted the subscript $j$ for the $j$-th output capsule for simplicity.} \emph{before} any routing
    or non-linearity as 
\begin{align}
    \label{eq:sabour_caps1}
    \vec s = \sum_i c_{i}\hat{\vec u}_{i}\,\,, \quad \hat{\vec u}_{i} = \mat{W}_{i} \vec u_i\quad\text{.}
\end{align}
Here, $\vec{u}_i$ is the capsule vector of capsule $i$ from the incoming layer and 
    $\mat W_{i}$ a matrix which transforms the capsule activations to generate the votes $\hat{\vec u}_{i}$,
    i.e., the vote of the $i$-th incoming capsule to the output capsule in the current layer.
    Note that $c_{i}$ are the coefficients for the dynamic routing algorithm. If no routing is applied, they
        can be combined with $\mat W_{i}$ to yield $\widetilde{\mat{W}}_{i} = c_{i}\mat W_{i}$,
    which simplifies eq.~\eqref{eq:sabour_caps1} to
\begin{align}
    \label{eq:sabour_caps2}
    \vec s = \sum_i \hat{\vec u}_{i}\,\,, \quad \hat{\vec u}_{i} = \widetilde{\mat{W}}_{i} \vec u_i\quad\text{.}
\end{align}
Further, we note that the linear transformation of $\vec u_i$ can be represented as votes of the individual
    entries $t$ of $\vec u_i$:
\begin{align}
    \label{eq:sabour_caps3}
    \hat{\vec u}_{i} = \widetilde{\mat{W}}_{i} \vec u_i = \sum_t \left[\vec u_i\right]_t \left[\widetilde{\mat W}_{i}^T\right]_t \quad\text{.}
\end{align}  
Here, 
    $\left[ \mat W\right]_t$ denotes the $t$-th row in a matrix $\mat W$ (equivalently for a vector).
Inserting this formulation of $\hat{\vec u}_{i}$ in eq.~\eqref{eq:sabour_caps2} yields
\begin{align}
    \label{eq:sabour_caps4}
    \vec s = \sum_i \sum_t \left[\vec u_i\right]_t \left[\widetilde{\mat W}_{i}^T\right]_t \quad\text{.}
\end{align}
Hence, $\vec s$ can be represented as the result of votes from all neurons $u$ contained in any of the incoming capsules (note that we sum over all entries in all capsules).
If we represent the activations of these neurons in a single vector $\vec x$, denote their activations by $x_u$,
    and their respective votes for the output capsule by ${\vec v}_u$
    we can write eq.~\eqref{eq:sabour_caps4} as:
\begin{align}
   \vec s = \sum_i \sum_t \displaystyle\overbrace{\left[\vec u_i\right]_t}^{\widehat{=} x_u} \;
                                \overbrace{\left[\widetilde{\mat W}_{i}^T\right]_t}^{\widehat{=} {\vec v}_u}
                                = \sum_u x_u {\vec v}_u
    \quad\text{.}
    \label{eq:sabour_caps6}
\end{align}
The sum on the right hand side in eq.~\eqref{eq:sabour_caps6}, in turn, can be expressed as a simple matrix-vector multiplication, such that
\begin{align}
    \vec s = \sum_u x_u\vec v_u = \mat v \vec x
    \label{eq:sabour_caps7}
\end{align}
with $\vec v_u$ as the columns of \tmat v.
The result is, of course, trivial and only shows that the capsule activations (a weighted sum of the inputs, eq.~\eqref{eq:sabour_caps1}) are obtained as a linear transformation of the input if no dynamic routing is applied.

Finally, we note that subsequent to this linear transformation, the authors in~\citesupp{sabour2017dynamic} apply the squashing function SQ (see eq.~(2) in our main paper) to the output vector $\vec s$, which yields the final capsule output $\text{CAP}(\vec x)$:
\begin{align}
    \text{CAP}(\vec x) = \text{SQ}(\vec s(\vec x)) = \text{SQ}(\mat v\vec x)\quad .
    \label{eq:sabour_caps8}
\end{align}
    
\subsection{Differences to capsules}
\label{subsec:capsule_diff_usage}
In the previous section we showed that the activation of a single capsule is computed as a linear transformation of all input activations, which is subsequently squashed (see eq.~\eqref{eq:sabour_caps8}).
Comparing this with the computation of the DAU outputs (eq.~(1) in our main paper), we see that this is equivalent to the dynamic weight computation if $\mat v =\mat{ab}$ and $\vec b=\vec 0$.
As such, Dynamic Alignment Units and capsules are related. However, there are important differences, which we discuss in the following and summarise in Table~\ref{tbl:caps}.

First, in~\citesupp{sabour2017dynamic}, the squashed capsule activations $\text{CAP}(\vec x)$
    are used as input to the next layer (after potentially applying the dynamic routing algorithm first).
    Instead of forwarding the squashed activations directly, in our DAUs they are used to linearly transform the input.
Second, by factorising the matrix $\mat v$ into two matrices $\mat{ab}$, we are able to control the rank of the linear transformation. Third,
when computing the weights in the DAUs we allow for an additional bias term, which in the context of capsules can be considered a `default vote'.
Fourth, we generalise the non-linearity in the DAUs to any non-linearity that only changes the norm.
Lastly, we do not apply dynamic routing in the DAUs.

\clearpage
{
\justifying
\bibliographystyleS{supplement/ieee_fullname}
\bibliographyS{supplement/supp_bib}
}

\end{document}